\documentclass[acmtog]{acmart}

\usepackage{booktabs}

\usepackage[ruled]{algorithm2e}
\usepackage{multicol}
\usepackage{multirow}
\usepackage{pifont}
\usepackage{xcolor}
\usepackage{colortbl}

\SetAlFnt{\small}
\SetAlCapFnt{\small}
\SetAlCapNameFnt{\small}
\SetAlCapHSkip{0pt}

\setcopyright{cc}
\setcctype{by}
\acmJournal{TOG}
\acmYear{2026}
\acmVolume{45}
\acmNumber{6}
\acmArticle{209}
\acmMonth{12}
\acmDOI{10.1145/3842500}

\begin{document}
\title{DynOLAT: OLAT Video Prediction for Controllable Dynamic Full-Body Relighting}
\author{Taoming Liu}
\authornote{Equal contribution.}
\email{taomingliu@zju.edu.cn}
\affiliation{
  \institution{State Key Lab of CAD\&CG, Zhejiang University}
  \city{Hangzhou}
  \country{China}
}

\author{Yinji ShenTu}
\authornotemark[1]
\email{styj@zju.edu.cn}
\affiliation{
  \institution{State Key Lab of CAD\&CG, Zhejiang University}
  \city{Hangzhou}
  \country{China}
}

\author{Guangzhao He}
\email{gh466@cornell.edu}
\affiliation{
  \institution{Cornell University}
  \city{New York}
  \country{United States of America}
}

\author{Yudong Jin}
\email{krahetx@gmail.com}
\affiliation{
  \institution{State Key Lab of CAD\&CG, Zhejiang University}
  \city{Hangzhou}
  \country{China}
}

\author{Yukang Ye}
\email{yeyukang@4dv.ai}
\affiliation{
  \institution{4DV.ai}
  \city{Hangzhou}
  \country{China}
}

\author{Siyu Zhang}
\email{derizhangzsy@gmail.com}
\affiliation{
  \institution{4DV.ai}
  \city{Hangzhou}
  \country{China}
}

\author{Fan Jia}
\email{jiafan.jf@antgroup.com}
\affiliation{
  \institution{Ant Group}
  \city{Hangzhou}
  \country{China}
}

\author{Wei Hua}
\email{huawei@zhejianglab.org}
\affiliation{
  \institution{Zhejiang University}
  \city{Hangzhou}
  \country{China}
}
\affiliation{
  \institution{Zhejiang Lab}
  \city{Hangzhou}
  \country{China}
}

\author{Hujun Bao}
\email{bao@cad.zju.edu.cn}
\affiliation{
  \institution{State Key Lab of CAD\&CG, Zhejiang University}
  \city{Hangzhou}
  \country{China}
}

\author{Xiaowei Zhou}
\authornote{Corresponding author.}
\email{xwzhou@zju.edu.cn}
\affiliation{
  \institution{State Key Lab of CAD\&CG, Zhejiang University}
  \city{Hangzhou}
  \country{China}
}

\begin{abstract}
Accurate lighting response is essential for realistic dynamic human relighting, because shadows, highlights, and color changes should follow the target illumination in both direction and relative strength.
Current learning-based relighting methods can generate visually plausible videos, but they often exhibit imprecise lighting responses: shadows and highlights may drift away from the target direction, and the output brightness may not faithfully follow the relative intensities in the target environment, making the result closer to appearance harmonization than controllable relighting.
To address this issue, we propose DynOLAT, a generalizable OLAT video prediction framework for dynamic full-body humans.
Instead of directly synthesizing a single relit appearance, our method predicts temporally consistent One-Light-At-a-Time (OLAT) basis videos from flat-lit inputs, and then linearly combines the predicted bases according to the target environment map.
This OLAT-based formulation makes the directional response explicit in the predicted bases, while the relative lighting strength is determined by the linear weights derived from the environment map.
We further introduce a dynamic capture pipeline and the first large-scale dynamic full-body OLAT dataset, comprising over 24 million frames from 224 identity-outfit combinations, where different outfits of the same person define separate combinations.
Experiments show that DynOLAT responds more faithfully to variations in target lighting direction and relative intensity than the tested baselines, enabling controllable video relighting while preserving fine appearance details and temporal coherence.
The dataset and processing tools will be publicly released to foster future research.
\end{abstract}



\begin{teaserfigure}
    \centering
    \includegraphics[width=\textwidth]{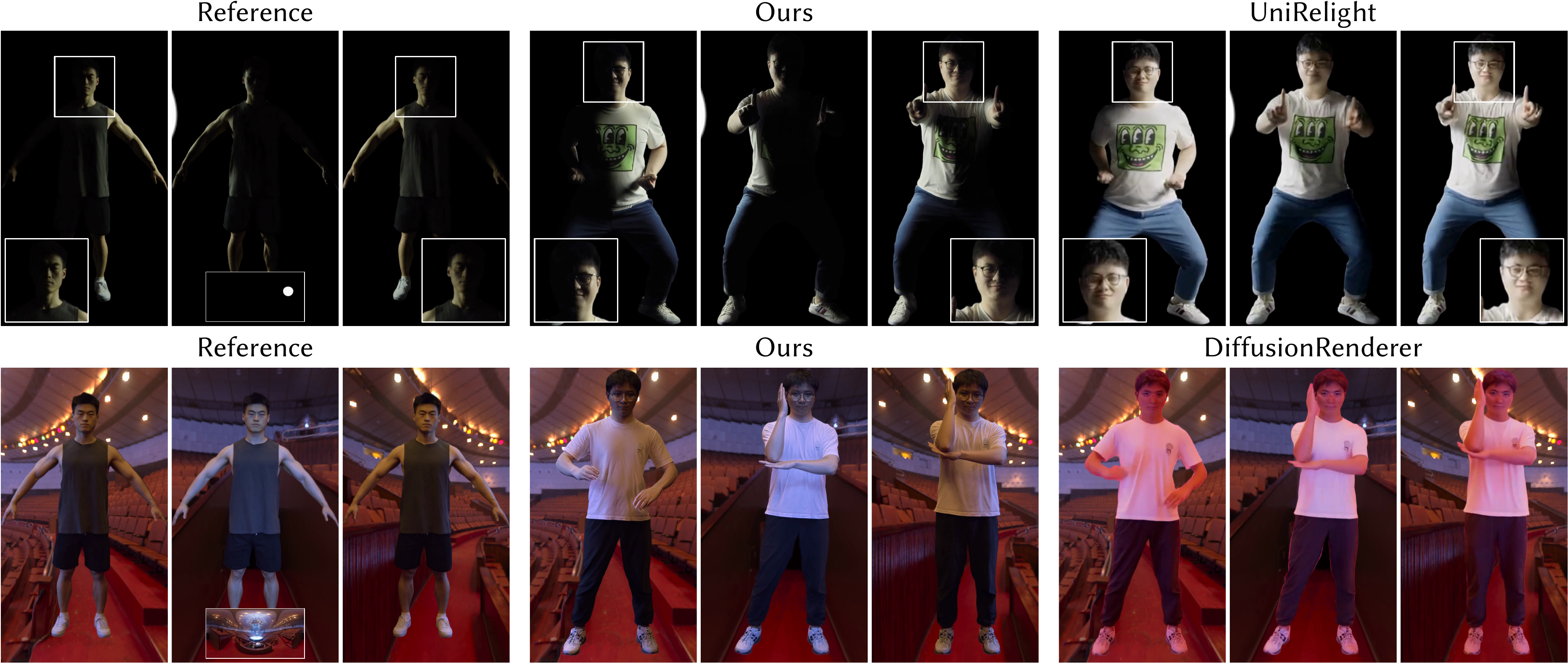}
    \caption{
    \textbf{Controllable dynamic full-body relighting from a single video.}
    Each row compares a lighting reference (left), ours (middle), and a baseline (right).
    Top: ours better matches the direction and shadow orientation of a rotating point light than UniRelight~\cite{he2025unirelight}.
    Bottom: ours follows relative environment-light intensity without the spurious red cast produced by DiffusionRenderer~\cite{DiffusionRenderer} in the low-intensity red-chair region.
    }
    \Description{Two-row comparison of lighting references, our results, and baseline results under rotating directional light and spatially varying environment illumination.}
    \label{fig:teaser}
\end{teaserfigure}

\maketitle

\section{Introduction}
Realistic dynamic human relighting requires the synthesized subject to respond accurately to the specified illumination in addition to appearing visually plausible. In practical VR/AR, film production, immersive telepresence, and game-content workflows, artists expect a relighting method to place shadows, highlights, and color changes according to the target light direction and relative intensity. This paper studies this controllability-oriented setting: given a monocular flat-lit video of a full-body human captured from an arbitrary viewpoint, we aim to synthesize a temporally coherent video of the same performance under a new lighting environment.

Existing general relighting methods still struggle with two forms of lighting-response accuracy that are critical for production use. First, their directional response is often imprecise: even when the target light direction is specified, the synthesized shading may exhibit a noticeable angular offset. Second, their intensity response is often unreliable: the output brightness may not faithfully follow the relative intensities in the target environment, making the result behave more like appearance harmonization than physically controlled relighting. These failure modes, highlighted in Fig.~\ref{fig:teaser} and Fig.~\ref{fig:total_comparison}, are not only local visual artifacts; they directly weaken control over the target light direction and relative strength.

The difficulty arises because monocular human relighting requires both appearance realism and disentangled light control. Traditional inverse-rendering methods~\cite{BarronTPAMI2015} attempt to estimate geometry, material, and illumination explicitly, but the problem is severely under-constrained for dynamic humans with spatially varying skin, hair, and clothing reflectance. Strong priors are therefore needed to resolve the ambiguity, often suppressing high-frequency details and producing overly smooth or artificial human appearance. Recent diffusion-based relighting methods~\cite{DiffusionRenderer, fang2025relightvid, bian2025relightmaster} reduce the need for explicit decomposition by learning direct image or video mappings from large-scale data. While these models can produce visually plausible results, their lighting conditions are often encoded as holistic environment images or prompts, making it difficult to guarantee that output shadows and intensities follow the target illumination in a directionally and radiometrically faithful way.

OLAT representations provide a natural route to controllable relighting because they separate light transport into directional basis responses. A one-light-at-a-time (OLAT) image records the subject response to a single lighting direction; arbitrary environment illumination can then be formed by linearly combining the corresponding bases. DifFRelight~\cite{he2024diffrelight} demonstrates that predicting OLAT representations can yield highly realistic and controllable facial relighting. Extending this idea from single facial images to dynamic full-body videos is non-trivial, because full-body motion introduces loose clothing deformation, self-occlusion, larger pose variation, and a lack of large-scale dynamic full-body OLAT supervision.

To address this gap, we propose DynOLAT, an OLAT video prediction framework for controllable dynamic full-body relighting. Given a flat-lit human video and a target lighting direction, our model predicts the corresponding OLAT basis video; the predicted bases are then combined in linear color space according to any target environment map. This formulation makes the target lighting explicit at two levels: the network learns directional basis responses, and the final relit video is assembled with environment-dependent intensity weights. As a result, the method is designed to preserve directional alignment and support explicit relative-intensity control, rather than relying on an unconstrained end-to-end generation of a single relit appearance.

Our technical design adapts a pretrained video model~\cite{wan2025} to this OLAT prediction task while preserving source identity, details, and temporal coherence. We adopt Latent Bridge Matching (LBM)~\cite{chadebec2025lbm} to transport flat-lit video latents directly toward OLAT latents, avoiding the source-detail loss introduced by sampling from independent Gaussian noise. We further inject the target lighting direction through block-specific AdaLN modulation, enabling fine-grained directional control across the video transformer. For long sequences, a recurrent bridge context reuses previously generated OLAT frames as analytic overlap constraints, maintaining coherence beyond the training window.

To train this data-driven model, we build a dedicated dynamic OLAT acquisition pipeline. The system rapidly alternates flat and directional illumination and then aligns the resulting frames to form paired flat-lit/OLAT videos. Using this pipeline, we collect a large-scale, multi-view, dynamic full-body OLAT dataset containing diverse subjects, clothing, motions, and viewpoints. To the best of our knowledge, this is the first dataset of comparable scale that provides dynamic full-body OLAT supervision for learning temporally consistent lighting responses. We will publicly release the dataset and processing tools to support future research.

Extensive experiments demonstrate that DynOLAT improves controllable full-body relighting in dynamic sequences. Qualitative comparisons show that our method better follows the target light direction and relative illumination strength, while preserving fine body details and temporal consistency. Quantitative comparisons and ablation studies further validate the effectiveness of our design.

In summary, our main contributions are as follows:
\begin{enumerate}
    \item An OLAT video prediction framework that formulates dynamic full-body relighting as directionally decomposed light-response synthesis, enabling more faithful responses to target light direction and environment-map intensity.
    \item Tailored model designs that adapt Latent Bridge Matching for source-detail preservation, use block-specific lighting modulation for directional control, and employ recurrent bridge context for long-sequence temporal coherence.
    \item The first large-scale dynamic full-body OLAT dataset with sufficient subject, clothing, motion, and viewpoint diversity for learning temporally consistent human light transport.
\end{enumerate}

\section{Related Work}

\subsection{Human-centric Relighting}
Relighting human performances from monocular video remains a fundamental challenge in computer graphics and vision. Existing methodologies can be broadly categorized into three paradigms: optimization-based inverse rendering, learning-based image relighting and learning-based video relighting.

\paragraph{Optimization-Based Inverse Rendering}
Leveraging recent advances in 3D neural representations like NeRF~\cite{mildenhall2021nerf} and 3DGS~\cite{kerbl20233d3DGS}, numerous works attempt to reconstruct human avatars and decompose intrinsic properties from multi-view images~\cite{li2025rgbavatar,hu2023gauhuman,BlendHeadAvatar,RTHumanAvatar,xu2024relightableAvatar,gao2024relightable,Relighting4D,chen2024meshavatarlearninghighqualitytriangular,singh2025relightableholoportedcharacterscapturing}.
By explicitly estimating geometry, material, and illumination, these methods provide a physically interpretable route to relighting control.
However, standard physically-based rendering (PBR) approximations often fail to capture complex surface effects, such as subsurface scattering and intricate self-shadowing, which affect both appearance realism and the accuracy of the predicted lighting response.
To circumvent these limitations, recent approaches utilize ground-truth OLAT scans to learn neural reflectance fields directly~\cite{li2024uravatar, zhou2023relightableUltraStage,fullbodyGaussianCodeAvatar,LightStageX,singh2025relightableholoportedcharacterscapturing}.
While these approaches have achieved photorealistic results for both upper-body~\cite{saito2024relightable, li2024uravatar} and full-body avatars~\cite{zhou2023relightableUltraStage, fullbodyGaussianCodeAvatar,LightStageX,singh2025relightableholoportedcharacterscapturing, teufel2026drex}, they rely heavily on subject-specific optimization.
This requirement for dense capture setups and per-identity training limits direct application to in-the-wild images or videos and makes controllable direction and intensity responses difficult for novel dynamic subjects.

\paragraph{Learning-Based Image Relighting}
To achieve generalization across subjects, a classical paradigm~\cite{nestmeyer2020learningphysicsguidedfacerelighting, pandey2021total, yeh2022learning, zhang2024physicallybasedface, ji2022relight, kim2024switchlight, tajimaEG25, mei2025lux} adapts the physics-based inverse rendering workflow into 2D feature space. These methods typically employ a cascaded pipeline that first decomposes the input image into intrinsic components (e.g., albedo and normals) and subsequently synthesizes the relit output via a neural renderer.
While this strategy explicitly models light transport, the multi-stage process is prone to error accumulation: inaccurate geometry or albedo can blur source details and also shift the resulting shadows, highlights, and brightness away from the target illumination.
Conversely, end-to-end approaches~\cite{sun2019sipr, dreamlight} map inputs directly to relit outputs.
Due to the scarcity of paired real-world data under varying illumination, many methods rely on synthetic data~\cite{ren2024relightful, ICLight} or auxiliary generation tasks~\cite{chaturvedi2025synthlight},
which can yield unrealistic shading or artifacts due to the domain gap.
More importantly for controllable relighting, direct output synthesis often uses a target environment or background as a holistic condition, providing limited constraints that ensure the generated shadow direction and relative brightness follow the specified illumination.
DifFRelight~\cite{he2024diffrelight} and POLAR~\cite{chen2025polar} fine-tune generative models on real OLAT data to predict OLATs for explicit lighting superposition, avoiding physical decomposition and synthetic-data domain gaps.
In contrast, 3DPR~\cite{rao2025threeDPR} predicts portrait OLATs with a 3D head prior and linearly combines them for relighting.
Nevertheless, applying these image-based solutions frame by frame to dynamic full-body sequences yields temporally unstable results, lacking the robustness required for video applications.

\begin{figure*}[t]
    \centering
    \includegraphics[width=\textwidth]{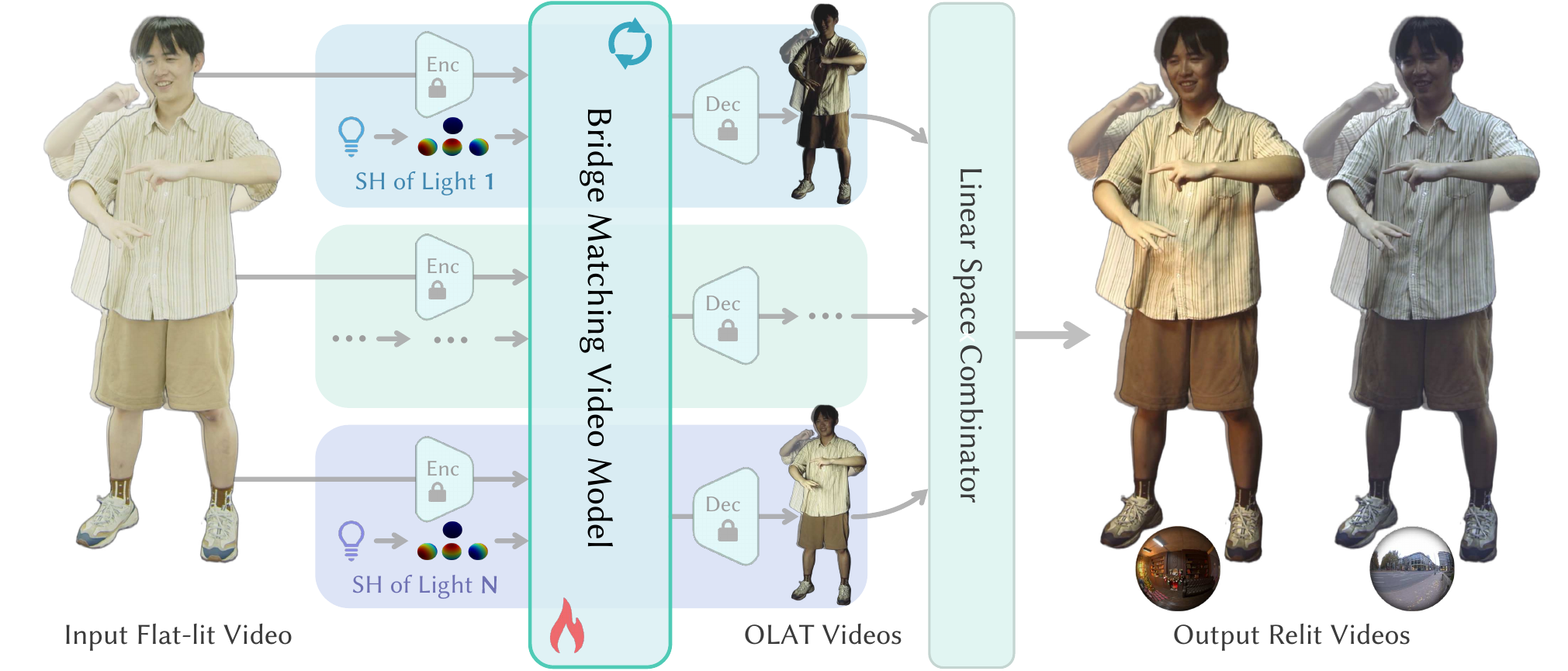}
    \caption{\textbf{The overview of the relighting pipeline.}
    Our model encodes a flat-lit video into latent features and uses Spherical Harmonics (SH) embeddings of the target light direction to modulate the generation. Through an iterative process, it transforms the flat-lit latents into target OLAT representations. These generated bases are then linearly combined to achieve high-fidelity relighting under any environment map.
    }
    \Description{Pipeline diagram showing flat-lit video encoding, target-light spherical-harmonic conditioning, iterative OLAT prediction, and linear combination under an environment map.}
    \label{fig:pipeline}
\end{figure*}

\paragraph{Learning-Based Video Relighting}
To extend relighting to videos, earlier methods attempted to enforce temporal consistency via loss constraints~\cite{DynamicOLAT} or feature blending~\cite{wang2025comprehensive}.
However, given the geometric and textural complexity of dynamic full-body motion, independent frame processing can struggle to maintain coherence at high resolution.
With the advent of video diffusion models~\cite{blattmann2023stablevideodiffusionscaling, videoworldsimulators2024, wan2025},
the mainstream paradigm has shifted toward fine-tuning video diffusion models~\cite{DiffusionRenderer, Guo_2025_CVPR, mei2025lux, lin2025illumicraft, fang2025relightvid, bian2025relightmaster, he2025unirelight, zeng2025lumen, liu2025uniLumos} or controlnets~\cite{zhang2025lumisculptenablingconsistentportrait}.
These methods typically inject condition frames by concatenating them with noisy latents, and represent lighting with environment maps, background images, or text-like conditions.
However, this standard conditioning strategy poses a fundamental conflict: the stochastic denoising process, while excellent for diversity, introduces randomness that undermines the deterministic nature of relighting. This often leads to hallucinated details or identity drift, where the generated texture deviates from the source, and can also weaken the alignment between the output illumination and the target direction or intensity distribution.
Instead of predicting a single relit video directly, our framework predicts OLAT basis videos and assembles the final illumination explicitly.
To make this basis prediction source-aligned, we employ bridge matching~\cite{chadebec2025lbm} from the input video latents to the target OLAT latents, preserving fine source textures while learning a more deterministic light-response mapping.
However, the success of such a data-driven video paradigm is heavily contingent on the availability of high-quality, temporally consistent training data.

\subsection{LightStage and OLAT Acquisition}
The LightStage architecture, pioneered by~\citet{debevec2000acquiring}, revolutionized reflectance acquisition through precise, programmable lighting control.
This setup inherently enables the capture of multi-view OLAT imagery.
Fundamentally, relying on the linearity of light transport, the appearance of a subject under arbitrary environmental illumination can be synthesized via the linear superposition of these OLAT basis images.
Consequently, LightStage data has become a "gold standard" benchmark for realistic relighting and has driven significant evolution in hardware design~\cite{LightStage3, LightStage4, LightStageX,he2024diffrelight}.

Despite its importance, the acquisition of such high-quality data remains heavily restricted by domain.
While numerous works have explored OLAT-based frameworks, they predominantly focus on restricted domains such as facial appearance~\cite{li2024uravatar,he2024diffrelight,chen2025polar}, hands~\cite{chen2024urhand}, or static objects~\cite{zhou2025olatverse, liu2023openillumination,toschi2023relight}. 
Research addressing full-body capture remains scarce~\cite{fullbodyGaussianCodeAvatar,teufelgera2025HumanOLAT,martinez2024codecStudio}, and publicly available datasets are extremely limited. 
More critically, prior work on time-multiplexed dynamic OLAT capture has focused on faces and upper bodies~\cite{LightStage3,DynamicOLAT}.

To the best of our knowledge, no existing public dataset provides dynamic full-body OLAT sequences at a comparable scale.
This data scarcity becomes a bottleneck for training generalizable full-body relighting models, which require a substantial diversity of human materials and motions to avoid overfitting.

To bridge this gap, we present our OLAT dataset, which includes a dedicated motion-rich subset. 
This component provides extensive dynamic OLAT data that covers diverse human materials and motions while supplying the directional light-response supervision needed to train controllable video relighting models.

\section{Method}

We decompose dynamic human relighting into explicit light-response prediction and linear environment assembly.
In the first stage, we employ our OLAT video prediction model to synthesize a set of basis OLAT videos from a flat-lit input sequence, where each basis corresponds to a specific lighting direction.
In the second stage, we reconstruct the final relit video under any complex environment map by linearly accumulating these predicted OLAT components in the linear color space.
This decomposition is central to our controllability goal: the lighting direction is handled by the predicted basis index, while the lighting intensity is handled by the environment-map weights.
An overview of our proposed framework is illustrated in Fig.~\ref{fig:pipeline}.

We first describe the architecture of our OLAT video prediction model in Sec.~\ref{subsec:olat_vid}.
We then explain how the flat-lit input frames (Sec.~\ref{subsec:frame_cond}) and the target lighting direction (Sec.~\ref{subsec:light_cond}) are processed and injected into the network.
Subsequently, we elaborate on our strategies for recovering high-frequency details (Sec.~\ref{subsec:high_freq}), which are necessary for producing reliable OLAT bases that retain the subject's material and texture response.
We then discuss our solution for inference on long video sequences (Sec.~\ref{subsec:long_vid}).
Finally, we present the image-based relighting (IBR) formulation that renders the final output via a linear combination of the predicted OLAT bases (Sec.~\ref{subsec:ibr}).

\subsection{Overview of OLAT Video Prediction Model}
\label{subsec:olat_vid}

Given a flat-lit video sequence and a target lighting direction, our goal is to synthesize the corresponding OLAT video.
Concretely, we operate within the compressed latent space of a pre-trained video diffusion model~\cite{wan2025}.
Adopting the formulation of~\citet{chadebec2025lbm}, we train the model to learn a flow trajectory that directly transports the source flat-lit latents to the target OLAT latents (see Sec.~\ref{subsec:frame_cond} for details).
During this process, the specific target lighting direction steers the generation by modulating the network features via Adaptive Layer Normalization (AdaLN)~\cite{perez2018film} (see Sec.~\ref{subsec:light_cond} for details).
Ultimately, the network predicts a velocity field to drive the transport of the input latents, which are subsequently decoded to yield the final high-fidelity OLAT video.

We explicitly model this process over video sequences rather than individual frames.
While static image relighting is well-established, the straightforward frame-by-frame application of such methods to dynamic sequences inherently suffers from temporal flickering and inconsistent shadowing.
By operating directly on video data, our model not only ensures temporal coherence and geometric consistency but also benefits from the diverse pose coverage naturally present in dynamic captures.

\subsection{Bridge Matching for Video Latents}
\label{subsec:frame_cond}
To predict a reliable OLAT basis video, it is crucial to feed the original video frames into the model in a way that preserves source appearance while supporting deterministic lighting edits.
Prior diffusion-based relighting methods typically concatenate input frames with Gaussian noise along a specific axis, such as the channel dimension~\cite{chaturvedi2025synthlight, mei2025lux} or the frame dimension~\cite{he2025unirelight, bian2025relightmaster}, to maintain the standard denoising process.
While this strategy supports the generative workflow, the reliance on randomly sampled noise often leads to outputs that fail to preserve fine textures and details from the input. Such inconsistency is unacceptable for human-centric relighting: texture drift changes the subject appearance, and the same stochasticity can also perturb the predicted shadows and highlights that should correspond to a specific lighting direction. See Fig.~\ref{fig:ablation_frame} for comparative results.

Since relighting is inherently a pixel-aligned editing task, we draw inspiration from LBM~\cite{chadebec2025lbm} and replace the input noise with latent tokens encoded from the flat-lit frames, allowing direct token-wise alignment between model inputs and outputs.
This source-to-target bridge helps preserve fine source textures while constraining the model to learn the lighting-dependent residual response needed for OLAT prediction.

Specifically, we implement this paradigm by constructing a stochastic bridge between the source and target distributions. We first encode the flat-lit video and the corresponding OLAT video into latent space via the encoder of a pretrained video variational autoencoder (VAE)~\cite{Kingma2013AutoEncodingVB}, producing latents $z_{\mathrm{flat}}$ and $z_{\mathrm{olat}}$. We then define the bridge interpolant $z_t$ as:
\begin{equation}\label{eq:latent_bridge}
    z_t = (1-t)z_{\mathrm{olat}} + tz_{\mathrm{flat}} + \sigma \sqrt{t(1-t)} \epsilon\,,
\end{equation}
where $t \in [0,1]$ is randomly sampled from a set of timesteps, $\epsilon \sim \mathcal{N}(0,I)$, and $\sigma$ denotes the noise strength, which we set to $0.001$ to enhance model robustness while keeping the input distribution within a reasonable range.
To learn the trajectory of this bridge, we regress a vector field using a video DiT model $v_{\Theta}$ by minimizing the bridge matching objective:
\begin{equation}\label{eq:latent_loss_function}
    \mathcal{L}_{\mathrm{LBM}} = \mathbb{E} \left[ \left\|\frac{(z_t - z_{\mathrm{olat}} )}{t} - v_{\Theta}(z_t, t) \right\|^2 \right]\,.
\end{equation}

At inference time, we feed the encoded flat-lit latents into the model and iteratively update them from $t=1$ toward the distribution of the corresponding OLAT latents via:
\begin{equation}\label{eq:iterative_sampling}
    \widehat{z}_{t+\Delta t} = z_t + v_{\Theta}(z_t,t) \Delta t\,,
\end{equation}
where $\Delta t < 0$, and the noise term is removed to maximally preserve information from the original input. When the iteration reaches $t=0$, we map the predicted $\widehat{z}_0$ back to the pixel space using the decoder of the video VAE.

\subsection{Lighting Direction Conditioning}
\label{subsec:light_cond}

To achieve precise directional control while maintaining a compact representation, we condition the model directly on the 3D lighting direction vector. This approach serves as a robust alternative to explicit spatial encodings (e.g., shading maps~\cite{pandey2021total, zeng2024dilightnet} or light source maps~\cite{choi2024scribblelight}), avoiding their associated computational overhead and potential geometric ambiguity.

\paragraph{Encoding}
To ensure geometric consistency with the visual content, we first rotate the world-space light vector into the camera coordinate system of the input video. Furthermore, to facilitate the learning of high-frequency illumination variations, we augment the raw direction vector $\mathrm{d}$ using Spherical Harmonics (SH) encoding $\mathcal{Y}(\cdot)$~\cite{tancik2023nerfstudio}. The final conditioning vector $c_{\mathrm{d}}$ is constructed by concatenating the raw vector with its positional encoding:
\begin{equation}\label{eq:lighting_condition}
    c_{\mathrm{d}} = \mathrm{d} \oplus \mathcal{Y}(\mathrm{d})\,,
\end{equation}
where $\oplus$ indicates concatenation along the channel dimension, and the SH degree is empirically set to $3$.

\paragraph{Injection}
We inject this lighting condition via an adaptive modulation mechanism. Within each transformer block, an individual linear projection layer maps the frequency-augmented vector $c_{\mathrm{d}}$ to the model’s hidden dimension.
This projected vector is subsequently added to the timestep embedding to modulate the input tokens by driving the scale, shift, and gate parameters of AdaLN~\cite{perez2018film}. To ensure the input distribution remains unperturbed at the onset of training, the parameters of these newly introduced linear layers are initialized to zero.

\subsection{High-Frequency Details Enhancement}
\label{subsec:high_freq}

Accurate OLAT prediction requires preserving high-frequency human appearance in addition to estimating the correct lighting response. Fine wrinkles, hair strands, fabric texture, and sharp shadow boundaries are part of the basis response that will later be linearly combined under environment illumination. On one hand, the iterative denoising process of diffusion models excels at producing visually realistic results~\cite{rombach2021highresolution, Nichol2021GLIDETP}, but the randomness introduced by noise can degrade high-frequency details present in the input. On the other hand, while a bridge matching approach in latent space naturally preserves much of the input pattern, it tends to excessively attenuate high-frequency details that should otherwise be preserved.
To make the predicted OLAT bases both visually faithful and reliable for controllable linear assembly, we strengthen high-frequency details across four complementary dimensions: image-video joint training, dynamic resolution, pixel-space supervision, and sampling iterations.

\paragraph{Image-Video Joint Training}
Although video data is essential for learning temporal dynamics, raw OLAT video captures frequently suffer from inherent quality degradations, such as motion blur, compression noise, and warping artifacts, which inevitably smooth out high-frequency details. In contrast, static OLAT imagery typically preserves superior textural fidelity and sharpness. To bridge this gap, we implement a joint training scheme that interleaves high-quality static images with video sequences. In this setup, static images guide the model to recover fine-grained structural and lighting details, while video data ensures temporal coherence.

\paragraph{Dynamic Resolution}
We also observe that full-body human figures in videos tend to be much taller than typical 16:9 frames, meaning that a fixed training resolution often compresses the subject’s effective pixels while leaving excessive background.
We therefore use no fixed training resolution: for each sequence, we crop all flat-lit and OLAT frames to the union of the flat-lit foreground masks predicted by~\citet{zheng2024birefnet}.
We then pad the crop, rather than resize the foreground, until both dimensions are divisible by 16 as required by the VAE.

\paragraph{Pixel-Space Supervision}
Supervision using only a latent-space regression loss is insufficient to achieve high visual quality, as it does not emphasize the preservation of high-frequency structures that are perceptually important, whereas a perceptual loss can better retain these visually salient patterns. 
This supervision also benefits controllable relighting because blurred basis predictions can smear shadow boundaries and weaken the directional cues used in the final composition.
Following~\citet{chadebec2025lbm}, we therefore incorporate a pixel-space LPIPS loss~\cite{zhang2018unreasonable}: the model’s predicted OLAT latents $\widehat{z}_{\mathrm{olat}}$ are decoded via the VAE into $\widehat{x}_{\mathrm{olat}}$ and an LPIPS loss is computed between $\widehat{x}_{\mathrm{olat}}$ and the ground truth $x_{\mathrm{olat}}$ in pixel space.
In practice, to avoid excessive memory usage on video data, we apply a random cropping strategy, computing the perceptual loss only on patches when the video resolution exceeds a predefined threshold. The overall training objective is then defined as:
\begin{equation}\label{eq:full_loss}
  \mathcal{L} = \mathcal{L}_{\mathrm{LBM}}(\mathcal{E}(x_{\mathrm{flat}}), \mathcal{E}(x_{\mathrm{olat}}), t) + \lambda\cdot\mathcal{L}_{\mathrm{perceptual}}(\widehat{x}_{\mathrm{olat}}, x_{\mathrm{olat}})\,,
\end{equation}
where $x_{\mathrm{flat}}$ and $x_{\mathrm{olat}}$ denote videos in pixel space, and $\mathcal{L}_{\mathrm{perceptual}}$ refers to the frame-wise LPIPS loss.

\paragraph{Sampling Iterations}
We find that simply reusing the single-step inference schedule from~\citet{chadebec2025lbm} leads to overly smooth outputs, where high-frequency details such as wrinkles on human skin and clothing are not well preserved. Performing iterative sampling with $4$ steps, however, sufficiently mitigates this issue and yields sharper OLAT bases for subsequent intensity-weighted composition.

\begin{figure*}[h]
    \centering
    \includegraphics[width=0.8\textwidth]{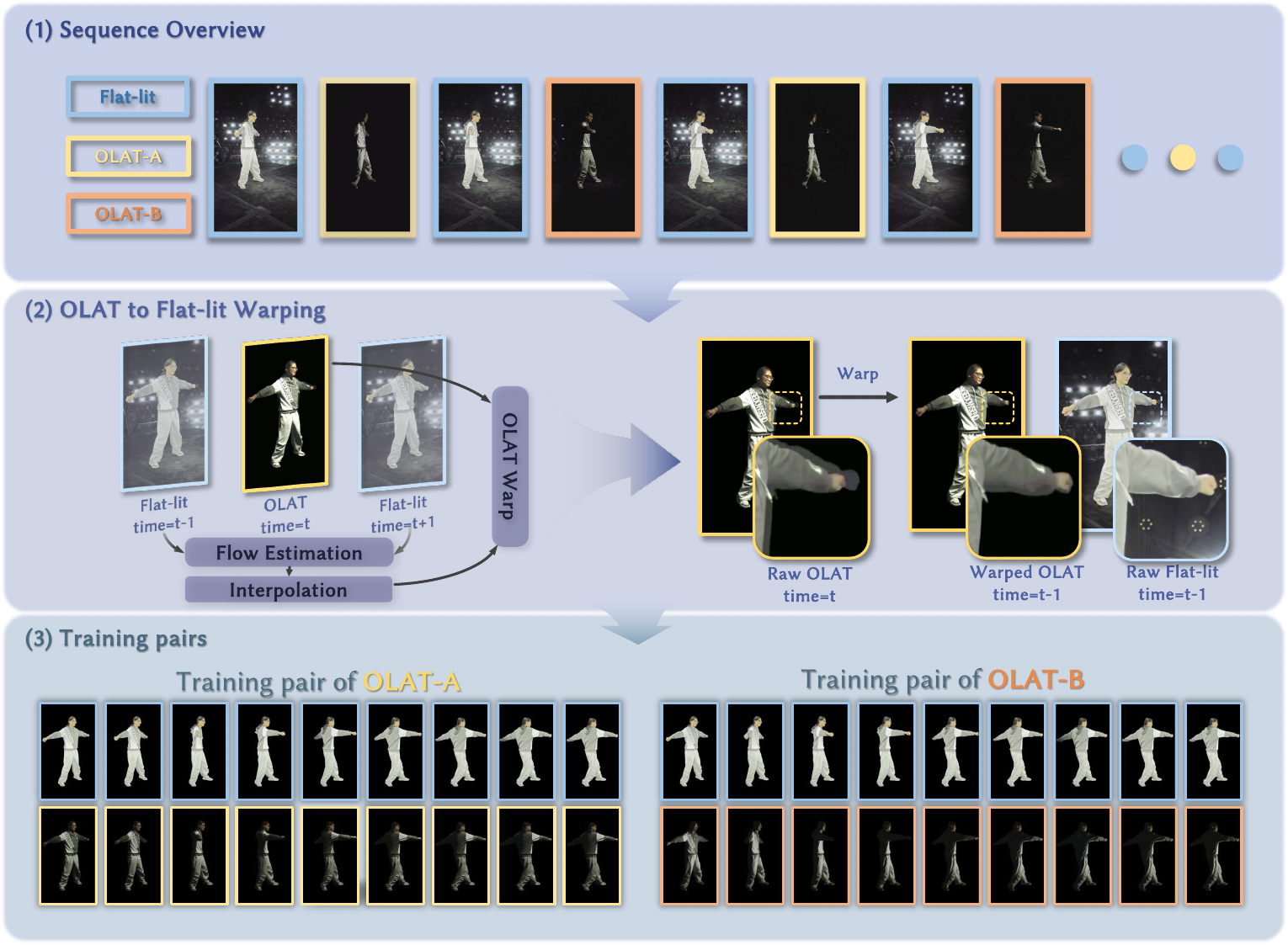}
    \caption{\textbf{Data processing pipeline for the \textit{Dynamic-subset}.}
    (1) Light Switch Sequence: One sequence adheres to a rigorous interleaved pattern: 18 flat-lit frames are paired with 9 frames of OLAT-A and 9 frames of OLAT-B, where A and B represent two distinct, spatially distant lighting directions.
    (2) OLAT to Flat-lit Warping: Within each sequence, we compute optical flow with AllTracker~\cite{harley2025alltracker} and foreground masks with BiRefNet~\cite{zheng2024birefnet} from the reference flat-lit frames. The OLAT frames are then warped to align with the flat-lit frames using the predicted flow to eliminate motion misalignment.
    (3) Training Pairs: Each sequence yields two temporally consistent video pairs. As shown, our light switching pattern successfully provides distinct illumination conditions for the identical motion sequence, ensuring robust disentanglement of light and geometry.
    For a detailed description of the processing pipeline, please refer to Sec.~\ref{subsec:data_processing}.
    }
    \Description{Three-stage data-processing diagram showing the interleaved light-switch sequence, optical-flow alignment of OLAT and flat-lit frames, and the resulting paired training videos.}
    \label{fig:data_process_pipeline}
\end{figure*}

\subsection{Long Sequence Inference}
\label{subsec:long_vid}
Video generative models are typically constrained to a fixed context length $F$ during training due to GPU memory constraints. Direct extrapolation to longer sequences is non-trivial; a naive approach of partitioning the video into independent chunks inevitably results in temporal discontinuities and flickering at the boundaries.
Existing sliding-window strategies~\cite{wu2024cat4d, jin2025diffuman4d} achieve consistency by fusing latents across varying Gaussian denoising levels. These are incompatible with our framework, where the generative trajectory is governed by a structured source distribution rather than stochastic noise.

To address this, we propose a recurrent bridge context, a strategy tailored for consistency in bridge-matching-based generation. We employ a sliding window with length $W=F$ and stride $S$, leveraging previously generated OLAT frames to guide the inference process. 
Specifically, we define a "history context" corresponding to the overlapping region of length $W-S$. This context is populated by $z_{\mathrm{olat}}^{\mathrm{prev}}\in\mathbb{R}^{(W-S)\times c\times h\times w}$, which denotes the finalized OLAT latents derived from the preceding window.
During the iterative sampling process of the current window, we enforce temporal continuity by overriding the estimated latents $\widehat{z}_{t}$ in this overlap region with the ideal bridge state derived from the history context:
\begin{equation}\label{eq:inference_replace}
    \widehat{z}_{t}[i] = 
    \begin{cases}
    (1-t)z_{\mathrm{olat}}^{\mathrm{prev}}[i] + tz_{\mathrm{flat}}[i], & \text{if }i < W-S, \\
    \widehat{z}_{t}[i], & \text{otherwise},
    \end{cases}
\end{equation}
where $z_{\mathrm{flat}}\in\mathbb{R}^{W\times c\times h\times w}$ represents the encoded flat-lit frames of the current window.
At each timestep, we replace the overlapping latents with the noise-free mean interpolant of Eq.~\ref{eq:latent_bridge}, thereby constraining the current window to match the previously generated frames in the overlap region and propagating temporal context forward.
Finally, we concatenate the context latents $z_{\mathrm{olat}}^{\mathrm{prev}}$ with the non-overlapping part of the predicted $\widehat{z}_{0}[W-S:W]\in\mathbb{R}^{S\times c\times h\times w}$ and decode the combined sequence via the video VAE.

This recursive strategy leverages the causal nature of the video VAE to extend bridge-matching inference beyond the training context while preserving consistent appearance and illumination responses without additional training.
Results on sequences of up to 917 frames are provided in the supplementary material.

\subsection{Linear Space Combination}
\label{subsec:ibr}

To relight the dynamic performance under a target high dynamic range (HDR) environment map, we associate each predicted OLAT basis with a light direction in the camera coordinate system. We partition the environment map according to the nearest basis direction and integrate the linear HDR radiance within each region to obtain its weight. The final video is synthesized by performing a frame-by-frame linear combination of the predicted OLAT videos. Specifically, for each frame, the corresponding outputs from all basis videos are converted to linear space, multiplied by these weights, and accumulated to produce the relit result. This linear-space assembly is the key step for intensity controllability: the relative strength of each environment region is explicitly determined by its combination weight, rather than being implicitly inferred by the model from an unconstrained conditioning image. A single global scaling factor, typically set to 3, is applied per environment map to match the dynamic range of the target illumination to the captured lighting bases, while preserving the relative weights among basis directions.

\section{The Dataset}
\label{sec:datasets}

\subsection{Overview}
\label{subsec:dataset_overview}

Training our controllable video relighting model requires large-scale paired video data consisting of a subject under flat lighting and the corresponding response to known OLAT illumination.

However, acquiring such ground truth is physically non-trivial, as a subject cannot simultaneously experience two distinct lighting conditions at the exact same instant. 
To address this, we employ a time-multiplexed acquisition protocol, rapidly alternating between flat-lit frames and OLAT frames. 
Since inevitable subject motion occurs during the interval between light switches, raw frames are inherently misaligned. We mitigate this by estimating optical flow between consecutive flat-lit frames and warping the intermediate OLAT frames to align spatially with the flat-lit references, thereby yielding pixel-aligned training pairs.

Crucially, to ensure the model learns robust representations, we devised a novel Light Switching Pattern.
Our empirical findings suggest that training on a single fixed light direction leads to overfitting, where the model fails to disentangle illumination from motion features.
To address this, we alternate between two distinct OLAT directions throughout a capture session. Specifically, we select the two lights with the greatest spatial distance within the LightStage to maximize illumination variance, while ensuring that all 101 distinct light sources are traversed exactly once across the full recording session of each subject, with no repetitions.
Each OLAT direction is repeated for 9 consecutive cycles to ensure sufficient temporal sampling. 
Consequently, these two sets of OLAT frames, along with their interleaved flat-lit references, constitute a block of 18 pairs, which we define as a single sequence.
The raw structure of such a sequence is visualized in Fig.~\ref{fig:data_process_pipeline}(1), while the processed, aligned pairs are shown in Fig.~\ref{fig:data_process_pipeline}(3).
Across a full recording session for one subject, this process yields 51 such sequences for each camera view.
Through this protocol, we acquired over 23 million valid frames across 189 identity-outfit combinations in the \textit{Dynamic-subset}.
We avoid alternating among too many directions (e.g., three or more) as it introduces excessive temporal gaps between frames of the same lighting condition, which degrades the temporal coherence of the video and hinders training.

Additionally, we acquired complete OLAT sequences from 35 identity-outfit combinations in static poses.
These dense captures support image-based relighting synthesis of ground-truth images under target environment maps, providing a controlled benchmark for image-level lighting evaluation.

Consequently, we organize our dataset into two primary components: the \textit{Dynamic-subset} (comprising the motion-rich paired video data) and the \textit{Static-subset} (comprising the comprehensive full OLAT sequences).
Tab.~\ref{tab:olat_dataset_compare} compares the scale and capture properties of our dataset with existing benchmarks. The supplementary material reports aggregate statistics for the 224 identity-outfit combinations and provides detailed dataset overviews.

All participants provided informed consent that explicitly permits research use and public release of their captured data.
The release will exclude unnecessary personal metadata and will document the intended research use, privacy considerations, and potential misuse in synthetic-media pipelines.
We plan to release the processed dataset upon publication under terms consistent with the participants' consent.

In the following subsections, we detail the technical implementation of our dataset construction. 
In Sec.~\ref{subsec:capture_setup}, we describe the Capture Setup, detailing the unified LightStage hardware and the distinct camera array configurations tailored for each subset. 
In Sec.~\ref{subsec:data_processing}, we present the data-processing pipeline, focusing on motion-artifact mitigation and data-quality assessment.

\begin{table*}[t]
\caption{
\textbf{Comparison of our OLAT dataset with existing OLAT datasets.}
To the best of our knowledge, this is the first dynamic full-body OLAT dataset.
It contains 24M valid frames, compared with 0.85M frames in the previous largest full-body OLAT dataset~\citep{teufelgera2025HumanOLAT}.
Column definitions:
\textbf{Type}: subject category (upper-body or full-body);
\textbf{Dynamic}: presence of dynamic OLAT motion sequences;
\textbf{\#Frames}: total frame count across all views;
\textbf{\#Subj.}: number of identity-outfit combinations;
\textbf{\#Views}: number of cameras, including short-lens and telephoto arrays;
\textbf{Res.}: approximate image resolution, where 1K $\approx$ $1920\times1080$, 4K $\approx$ $3840\times2160$, and 8K $\approx$ $7680\times4320$.}
\label{tab:olat_dataset_compare}
\begin{center}
\resizebox{0.85\textwidth}{!}{
\begin{tabular}{l c c c c c c}
\toprule
 Dataset & Type & Dynamic & \#Frames & \#Subj. & \#Views & Res. \\
  \midrule
  Dynamic OLAT~\cite{DynamicOLAT} & upper-body & \ding{51} & $603$k & $4$ & $1$ &  $4$K \\ 
  Goliath-4~\cite{martinez2024codecStudio} & upper-body & \ding{51} & ${>}1$M & $4$ & $110$ &   $4$K \\
  NetFlix Data~\cite{mei2025lux} & upper-body & \ding{51} & $12$M & $67$ & $36$ & $4$K \\ 
  POLAR~\cite{chen2025polar} & upper-body & \ding{51} & \textbf{28M} & $220$ & $32$ & $4$K \\ 
  \midrule
  Ultrastage~\cite{zhou2023relightableUltraStage} & full-body & \ding{55} & $192$k & $100$ & $32$ & $8$K \\
  HumanOLAT~\cite{teufelgera2025HumanOLAT} & full-body & \ding{55} & $850$k & $21$ & $40$ & $5$K \\
  \midrule
  \textit{Ours (Static-subset)} & full-body & \ding{55} & $1.06$M & $35$ & $70$ & $4$K \\
  \textit{Ours (Dynamic-subset)} & full-body & \ding{51} & \textbf{23M} & $189$ & $68$ & $4$K \\
  \textit{Ours (Total OLAT)} & full-body & \ding{51} &\textbf{24M} & $224$ & $68/70$ & $4$K \\
 \bottomrule
\end{tabular}
}
\end{center}
\vspace{-4mm}
\end{table*}

\subsection{Capture Setup}
\label{subsec:capture_setup}

\begin{figure}
    \centering
    \includegraphics[width=\linewidth]{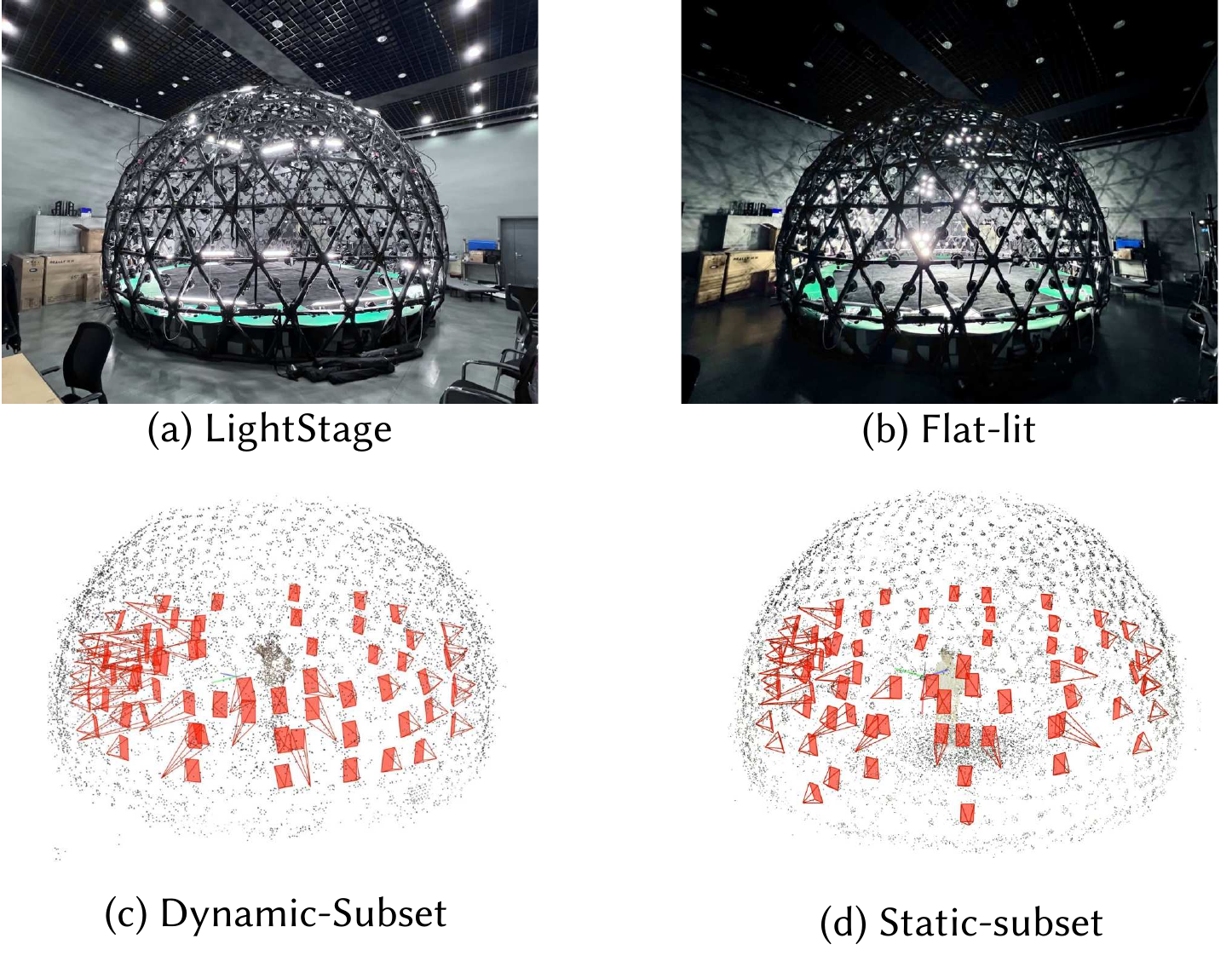}
    \caption{\textbf{Capture system overview.}
    (a) Our LightStage.
    (b) The flat-lit illumination configuration.
    (c--d) Camera distributions for the \textit{Dynamic-subset} (c) and the \textit{Static-subset} (d), reconstructed using COLMAP~\cite{colmap}.}
    \Description{Photographs of the LightStage and flat-lit configuration, followed by reconstructed camera layouts for the Dynamic- and Static-subsets.}
    \label{fig:cam_env_overview}
\end{figure}

\paragraph{LightStage System.}
Our illumination system is shared across both subsets (Fig.~\ref{fig:cam_env_overview}(a)).
It consists of over 800 LEDs grouped into 101 finite-area, individually addressable lighting units (referred to as One-Light units), driven by a custom microcontroller with microsecond precision.
For dynamic acquisition, we configure the lighting clock cycle to 35 ms.

\paragraph{Camera Systems and Settings.}
We deployed two synchronized camera systems, one for each subset.
We prioritized storage efficiency and bandwidth for the \textit{Dynamic-subset} to enable large-scale recording, while prioritizing dynamic range and raw quality for the \textit{Static-subset}.

\begin{itemize}
    \item \textbf{\textit{Dynamic-subset}:} We used 68 custom-built PTZ cameras (Sony IMX678) to record 4K H.264 video at 60 fps (60 Mbps bitrate, I-frame interval of 2).
    The array comprises 48 \textbf{short-lens} cameras (approximately 9 mm) distributed spherically and 20 \textbf{telephoto} cameras (approximately 34 mm) placed primarily in frontal positions to capture upper-body details.
    Exposure is fixed at $1/125$ s (f/4.8) to reduce motion blur.
    
    \item \textbf{\textit{Static-subset}:} We utilized 70 Blackmagic Micro Studio Camera 4K G2 units arranged in a similar spherical layout: 54 units (14/15 mm lenses, including six zenith/nadir cameras) and 16 telephoto units (47 mm lenses).
    The raw footage is stored as Blackmagic RAW. While this format provides superior fidelity, its prohibitive bitrate prevents further scaling.
    We use a faster shutter speed of $1/500$ s (f/4.4, ISO 3200) to reduce motion blur and transcode the footage to 100 Mbps for processing.
\end{itemize}
The spatial distributions are visualized in Fig.~\ref{fig:cam_env_overview}(c) and (d). 
Representative images from these camera arrays are provided in the supplementary material.

\paragraph{Illumination Strategy.}
A major challenge in OLAT capture is the extreme dynamic range discrepancy between full-on lighting and single OLAT conditions.
To address this discrepancy, we activate a uniformly sampled subset (1/6) of the lights to simulate ambient illumination (Fig.~\ref{fig:cam_env_overview}(b)).
This preserves perceptual consistency within the camera's limited dynamic range. Note that all references to "flat-lit" in this paper refer to this flat-lit configuration.

\paragraph{Light Switching Patterns.}
The two subsets employ different switching patterns to serve their respective goals:
\begin{itemize}
    \item \textbf{\textit{Dynamic-subset} Pattern:} As introduced in Sec.~\ref{subsec:dataset_overview}, we repeat one OLAT direction for 9 cycles, interleaved with its spatially most distant counterpart for another 9 cycles (Fig.~\ref{fig:data_process_pipeline}(3)). This variation encourages the model to decouple motion from lighting.
    
    \item \textbf{\textit{Static-subset} Pattern:} We use 10 consecutive OLAT frames followed by 1 flat-lit frame. 
    For each identity-outfit combination, we capture at least three static poses and several flat-lit dynamic sequences for relighting evaluation.
\end{itemize}

\subsection{Data Processing}
\label{subsec:data_processing}

Given the complexity of massive-scale OLAT acquisition, especially for the \textit{Dynamic-subset}, the raw data requires careful processing to achieve geometric alignment and photometric consistency.
The overall pipeline is illustrated in Fig.~\ref{fig:data_process_pipeline}.
In the following sections, we detail the core algorithms for OLAT to flat-lit warping, followed by specific strategies for data quality assessment.

\paragraph{OLAT to Flat-lit Warping.}

We use optical flow to compensate for subject motion, including rapid actions in the \textit{Dynamic-subset} and involuntary sway in the \textit{Static-subset}, that causes misalignment between flat-lit and OLAT frames.
We compute bidirectional optical flow between consecutive flat-lit frames~\cite{harley2025alltracker}.
Assuming linear motion within the short interval, we derive the flow field for the intermediate OLAT frame via temporal linear interpolation.
Alongside flow estimation, we generate foreground masks for the flat-lit frames using BiRefNet~\cite{zheng2024birefnet} and compute a unified bounding box.

Based on the computed flow, we apply different warping strategies tailored to the downstream application:
\begin{itemize}
    \item \textbf{Training Pairs (Dynamic \& Static):} To minimize warping artifacts, we warp each OLAT frame to its temporal nearest-neighbor flat-lit frame. This yields diverse, well-aligned pairs for learning-based models, preserving the original motion fidelity.
    \item \textbf{Quality Assessment Stack (Static only):} For quality evaluation, we warp all OLAT frames within a pose to the first flat-lit frame of the sequence, producing a geometrically aligned OLAT stack for the photometric analyses described below.
\end{itemize}

\begin{figure}
    \centering
    \includegraphics[width=\linewidth]{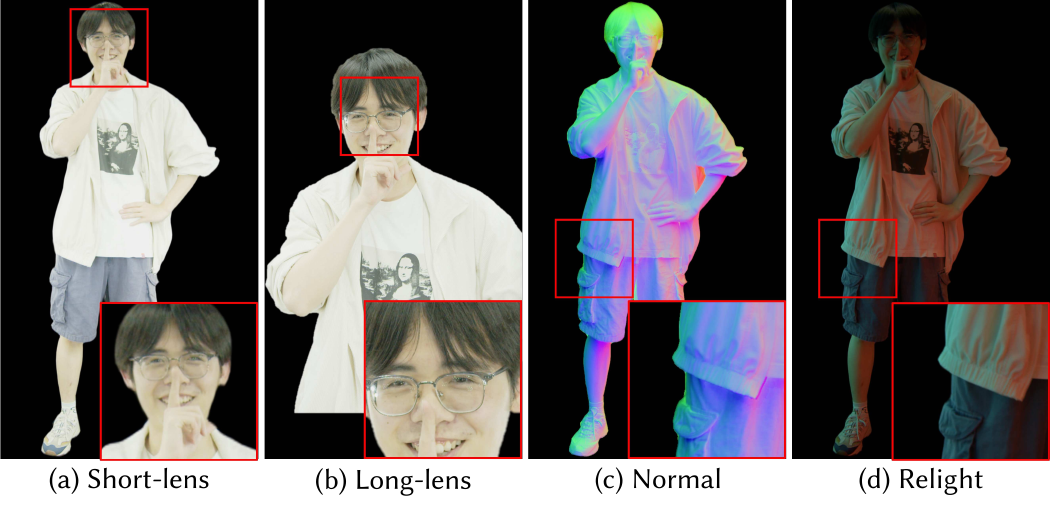}
    \caption{
        \textbf{Quality assessment.}
            (a) Full-body view from the short-lens array.
            (b) High-resolution view from the telephoto array, capturing fine facial details.
            (c) Surface normals estimated via photometric stereo from the full OLAT stack, recovering pore-level geometry.
            (d) Image-based relighting result synthesized via linear superposition, illustrating the photometric consistency and dynamic range of our data.
        }
    \Description{Full-body and telephoto captures, estimated surface normals, and an image-based relighting result from the captured OLAT stack.}
    \label{fig:BMD_validation}
\end{figure}

\paragraph{Quality Assessment.}
We utilize the geometrically aligned OLAT stack from the \textit{Static-subset} to quantitatively assess the quality of our acquisition system.
As illustrated in Fig.~\ref{fig:BMD_validation}, our raw captures from both short-lens (a) and telephoto (b) cameras preserve high-frequency details.
To assess geometric detail, we estimate surface normals via photometric stereo using gradient-descent optimization with outlier rejection. The resulting normal maps in Fig.~\ref{fig:BMD_validation}(c) retain fine-scale skin texture and clothing wrinkles.
To assess photometric linearity, we synthesize relit images by linearly combining the OLAT bases with environment-map weights after gamma correction.
The synthesized results (d) preserve fine shading and self-shadowing details more faithfully than the evaluated neural relighting baselines.

\section{Experiments}

\subsection{Implementation Details}

\begin{figure*}[t]
    \centering
    \includegraphics[width=\linewidth]{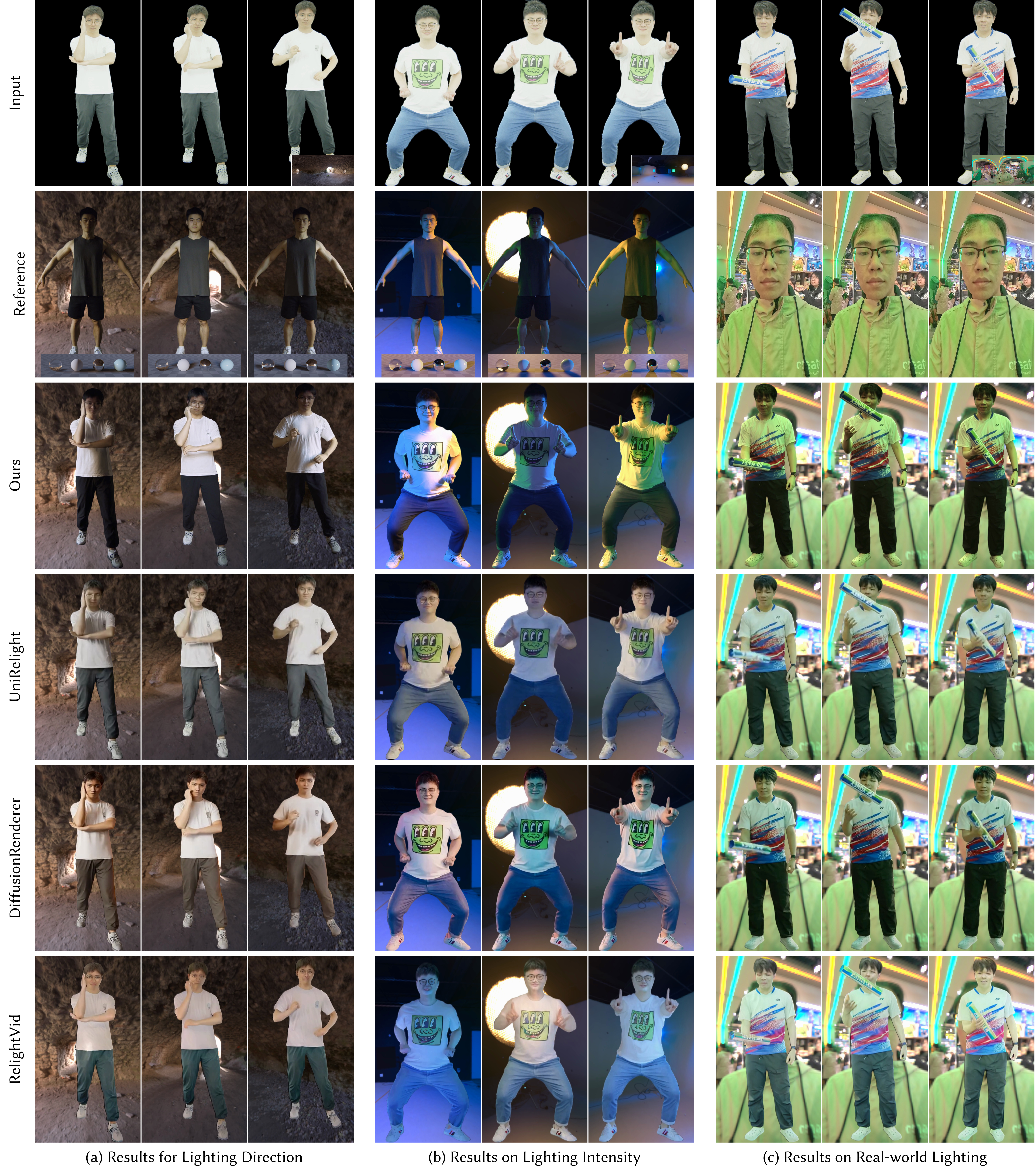}
    \vspace{-4mm}
    \caption{\textbf{Qualitative comparisons for video relighting.}
    We present results on our OLAT dataset under various environment maps.
    }
    \Description{Video-relighting comparisons among the input, target-light reference, our method, and baseline methods under several environment maps.}
    \label{fig:total_comparison}
    \vspace{-4mm}
\end{figure*}

\begin{figure}
    \centering
    \includegraphics[width=\linewidth]{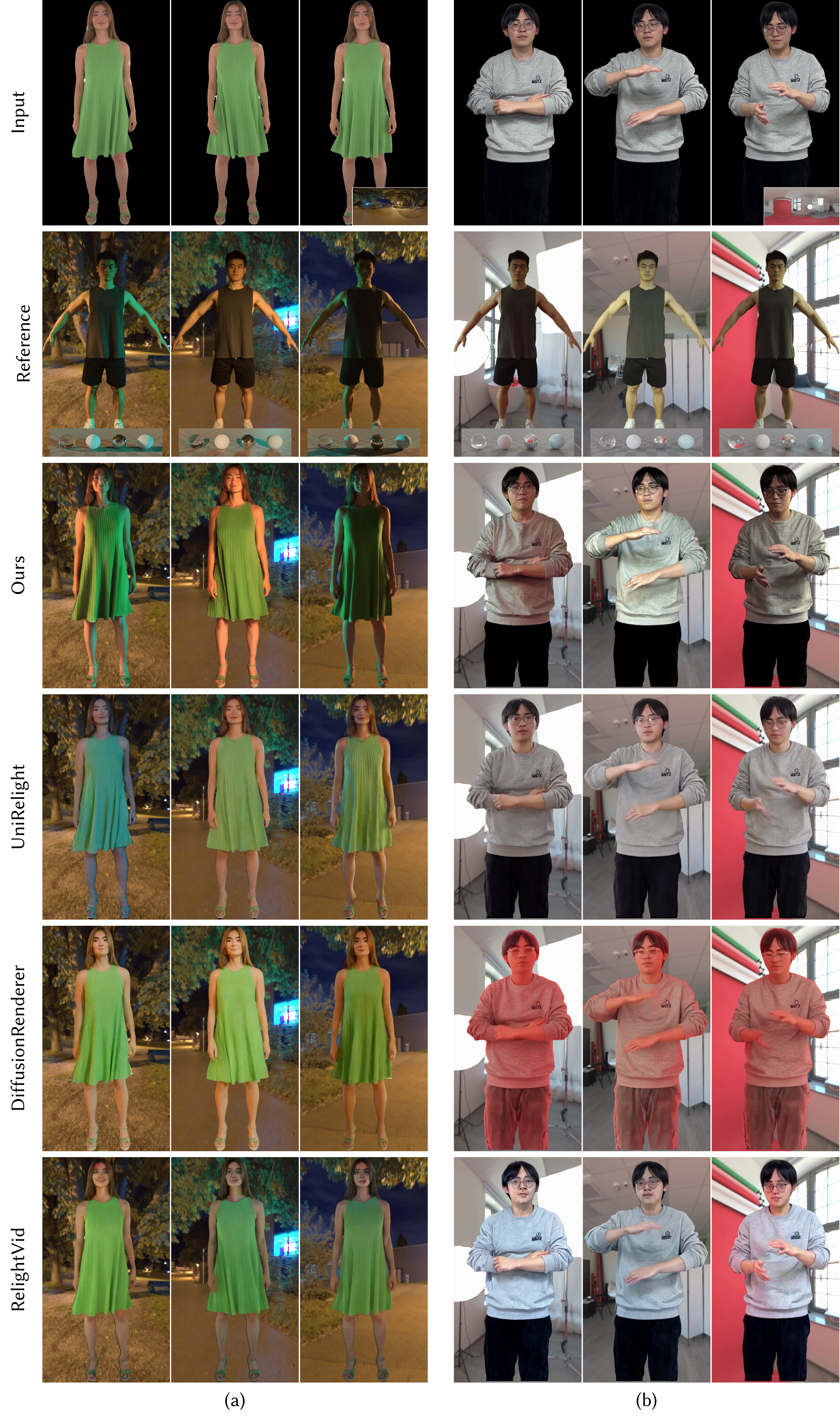}
    \vspace{-4mm}
    \caption{\textbf{Qualitative comparisons for in-the-wild video relighting.}
    We evaluate our method on (a) the Actor-HQ dataset~\cite{isik2023humanrfActorHQ} and (b) unstructured mobile phone captures.
    }
    \Description{Relighting outputs from our method and baselines for Actor-HQ and mobile-phone video inputs.}
    \label{fig:wild_comparison}
    \vspace{-4mm}
\end{figure}

\begin{figure*}[h]
    \centering
    \includegraphics[width=\linewidth]{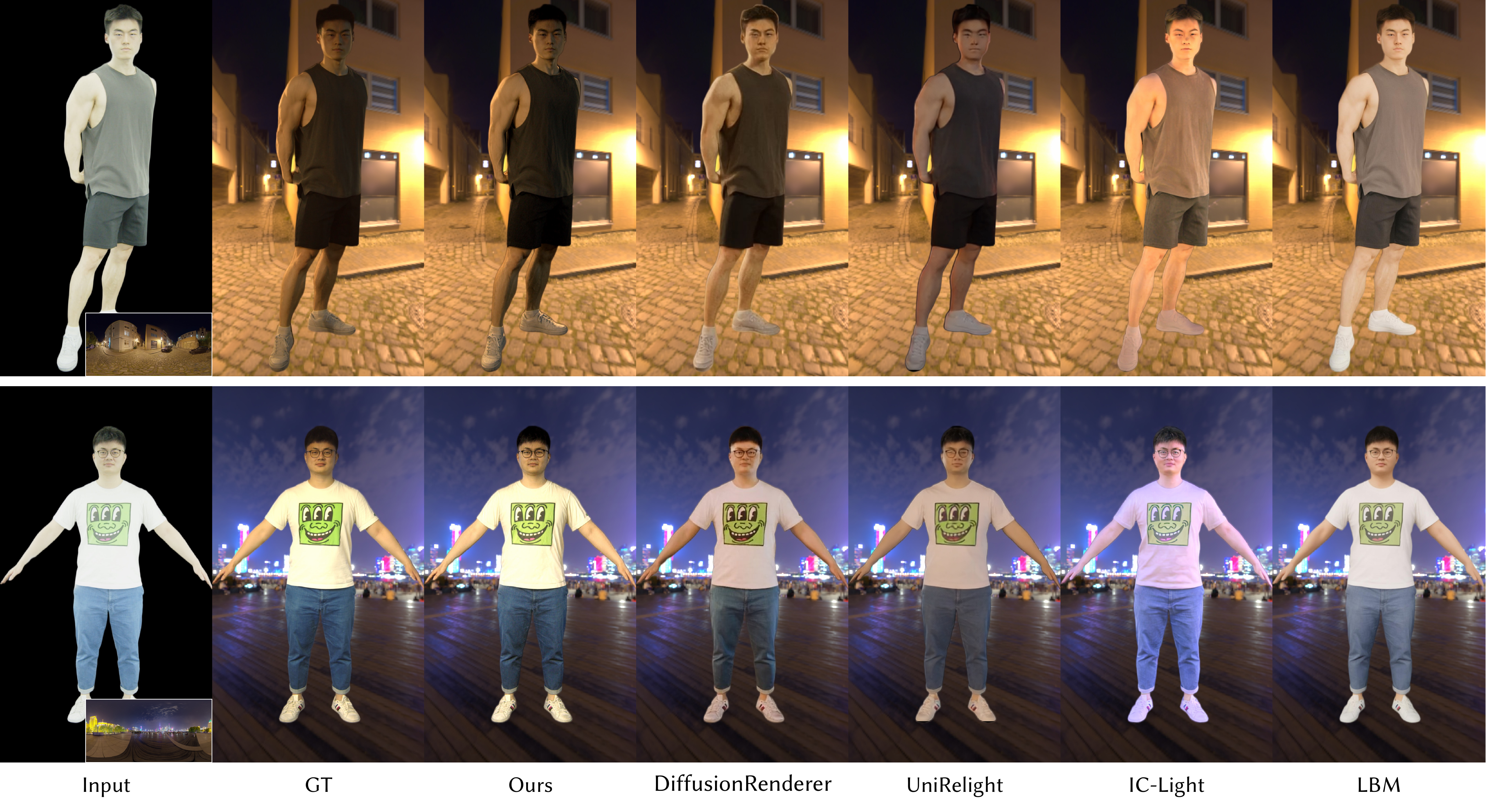}
    \vspace{-8mm}
    \caption{\textbf{Qualitative comparisons for image relighting.}
    Compared with the evaluated baselines, our result retains more of the input's fine-grained structure and more closely matches the captured OLAT-based ground truth.
    The top row highlights the specular response on the face and the shading contrast along the subject's rear arm.
    }
    \Description{Image-relighting comparison with captured OLAT-based ground truth, emphasizing facial highlights, arm shading, and fine subject details.}
    \label{fig:image_comparison}
    \vspace{-4mm}
\end{figure*}

\paragraph{Training Configuration}
We perform full-parameter fine-tuning based on the Wan2.1 1.3B foundation model~\cite{wan2025}. The model is trained for a total of 60,000 steps with a constant learning rate of $10^{-4}$ and a batch size of $8$. 
During training, the target OLAT directions are randomly sampled from all available lighting directions.
The probability of sampling static images for joint training follows a stepwise curriculum: it is set to $0.1$, $0.2$, and $0.3$ for each consecutive 20,000-step interval.
We use the same few-timestep sampling during training as~\citet{chadebec2025lbm}, where the discrete timesteps are sampled uniformly from the set $\{\mathrm{i}/16\}_{\mathrm{i}=1}^{16}$.
For the objective function, we utilize a VGG16-based LPIPS metric for the perceptual loss, setting the weight coefficient $\lambda$ to $1$.

\paragraph{Fixed Text Prompt}
Throughout both training and inference, we utilize a fixed, descriptive text prompt to provide a consistent semantic context:
\textit{"Given: a reference video of a person and a lighting direction representing a specific point light source. Generate: a realistic relit video of the same person under the target lighting defined by one light. Requirements: preserve identity and appearance, maintain consistent clothing and proportions, correct perspective, realistic lighting, shadows, and no artifacts."}

\paragraph{Inference Configuration}
During inference, we disable classifier-free guidance to rely primarily on our conditioning signals.
To ensure temporal consistency, the recurrent bridge context is configured with a window length of $W=9$ frames and a stride of $S=4$, which is essential for stable video generation.
We employ a uniform sampling strategy with 4 discrete timesteps, specifically $\{1,0.75,0.5,0.25\}$.
For each input sequence, we typically generate a set of 201 distinct OLAT videos to serve as the bases for relighting. 
These 201 bases preserve the zenith direction of the original 101-light configuration while doubling the azimuthal sampling density at each elevation.
Additional results in the supplementary material show comparable quality under static lighting but smoother responses to rotating high-frequency illumination with 201 bases; 101 bases reduce generation and storage cost when finer sampling is unnecessary.

\paragraph{Hardware and Runtime}
The entire training process is conducted on 8 NVIDIA A100 GPUs and takes approximately one week to complete.
We additionally measure the inference cost of generating the OLAT basis set.
For a 57-frame video at $1920\times1088$, generating one OLAT basis takes 119.45 seconds. When generation is distributed across eight A100 GPUs, producing 101 bases takes 26.89 minutes, while producing 201 bases takes 52.01 minutes.
At $1280\times720$, generating one basis takes 61.08 seconds; with the same eight-GPU setup, producing 101 bases takes 14.53 minutes, while producing 201 bases takes 27.56 minutes.
Because individual bases can be generated independently, they can be computed in parallel and cached for subsequent linear combinations without online network inference, although generation and storage costs scale approximately with their number.

\subsection{Datasets and Metrics}

We construct our training set by randomly selecting 100 identity-outfit combinations from the \textit{Dynamic-subset} and 28 from the \textit{Static-subset}.
For evaluation, we hold out 54 and 7 identity-outfit combinations from the two subsets, respectively.
For the \textit{Dynamic-subset}, we use only views captured with short-lens cameras.

To ensure alignment with the pre-training distribution of the foundation model, all data are represented in the sRGB color space. Additionally, to isolate the subjects and eliminate background interference, we apply a uniform mask to all input frames, setting the background to black.

Following the evaluation protocol of~\citet{he2024diffrelight}, we employ PSNR, SSIM, LPIPS~\cite{zhang2018unreasonable} (with an AlexNet backbone), and FLIP~\cite{andersson2020flip} to quantitatively assess model performance.

\subsection{Comparisons with Existing Methods}
We compare our model with three video relighting approaches: the open-source RelightVid~\cite{fang2025relightvid}, the Cosmos-based DiffusionRenderer~\cite{DiffusionRenderer}, and UniRelight~\cite{he2025unirelight}.
To establish a comprehensive benchmark, we also include two recent single-image relighting methods: IC-Light~\cite{ICLight} and LBM~\cite{chadebec2025lbm}.
For the open-source baselines, we resize input videos or images to their recommended resolutions and adapt the lighting inputs to each method's requirements. For DiffusionRenderer and UniRelight, we use the full HDR environment map as the condition. For RelightVid, IC-Light, and LBM, which rely on a background image, we sample the corresponding field of view from the HDR map and convert it to sRGB.

We qualitatively evaluate video relighting on our OLAT dataset and in-the-wild sequences in Figs.~\ref{fig:total_comparison} and~\ref{fig:wild_comparison}.
Reference images are generated via image-based relighting from captured OLAT images, together with material balls rendered under the target environment maps.

Fig.~\ref{fig:total_comparison} qualitatively compares responses to target direction (a), intensity distributions (b), and real-world lighting (c).
Our results follow the target changes more closely than the evaluated baselines.
The supplementary material further shows how our method responds to different intensity scalings of the same environment map.

To reduce the domain gap between the required flat-lit input and in-the-wild videos, we apply the delighting model described in the supplementary material before OLAT prediction.
Consequently, beyond controlled setups, Fig.~\ref{fig:wild_comparison} highlights our strong generalization capabilities to novel, in-the-wild scenarios.
Our model robustly preserves fine details and subject identity far better than the baselines.
This superiority is evident in the fine-grained environmental interactions it synthesizes, such as the physically precise shadows in (a) and the accurate reflected bounce light from the background red curtain in the first frame of (b).
Overall, these results confirm that our method synthesizes more photorealistic appearances and physically plausible lighting effects compared to existing approaches.

For quantitative lighting evaluation, we use the \textit{Static-subset}, whose complete captured OLAT sequences allow us to synthesize ground-truth images under target environment maps via image-based relighting. In this setting, we focus our comparison on two representative image-based methods (IC-Light~\cite{ICLight} and LBM~\cite{chadebec2025lbm}) and two video-based methods (DiffusionRenderer~\cite{DiffusionRenderer} and UniRelight~\cite{he2025unirelight}).

Image-relighting results are reported in Tab.~\ref{tab:image_comparison} and Fig.~\ref{fig:image_comparison}, with an additional qualitative HumanOLAT comparison provided in the supplementary material.
We resize all outputs and ground truth to the minimum evaluated resolution and mask their backgrounds.
Following~\citet{jin2024neural_gaffer}, each predicted color channel is rescaled to the ground-truth mean to resolve global illumination--surface-color scale ambiguity.
This calibration is used in quantitative evaluation to ensure a consistent comparison across methods. The results demonstrate that our method outperforms the baselines across most metrics and preserves significantly richer fine-grained details.

\begin{table}[h]
\caption{\textbf{Quantitative comparison on image relighting.}
To ensure a fair assessment, we employ a channel-aligned evaluation strategy to mitigate the inherent scale ambiguity between lighting intensity and surface color.
Under this protocol, our method better matches the captured OLAT-based relighting ground truth across most metrics, indicating improved light-response fidelity.}
\label{tab:image_comparison}
\begin{center}
\resizebox{\linewidth}{!}{
\begin{tabular}{l c c c c }
\toprule
  Methods& PSNR$\uparrow$ & SSIM$\uparrow$ & LPIPS$\downarrow$ & FLIP$\downarrow$ \\ 
  \midrule
  IC-Light~\cite{ICLight} & 19.51& .8760& .1322& .1226\\ 
  LBM~\cite{chadebec2025lbm} & 23.08&  \cellcolor{red!15}.9185& .1002& .0874\\
  UniRelight~\cite{he2025unirelight} & 22.91& .9070& .0928& .0889\\
  DiffusionRenderer~\cite{DiffusionRenderer} & \cellcolor{yellow!15}23.11& .9057& \cellcolor{yellow!15}.0900& \cellcolor{yellow!15}.0841\\
  \midrule
  Ours & \cellcolor{red!15}26.23& \cellcolor{yellow!15}.9169& \cellcolor{red!15}.0567& \cellcolor{red!15}.0602\\ \bottomrule
\end{tabular}
}
\par
\footnotesize
\colorbox{red!15}{\rule{0pt}{1em}\hspace{1em}} Best result \hspace{1em} \colorbox{yellow!15}{\rule{0pt}{1em}\hspace{1em}} Second-best result
\end{center}
\vspace{-2mm}
\end{table}

Following Lux Post Facto~\cite{mei2025lux}, we evaluate temporal consistency using LPIPS between adjacent frames and warping error (WE) computed with optical flow from RAFT~\cite{teed2020raftrecurrentallpairsfield} and AllTracker~\cite{harley2025alltracker}.
Tab.~\ref{tab:temporal_consistency} shows that our method achieves excellent temporal consistency.

\begin{table}[h]
\caption{\textbf{Quantitative comparison of temporal consistency.}}
\label{tab:temporal_consistency}
\begin{center}
\resizebox{\linewidth}{!}{
\begin{tabular}{l c c c c }
\toprule
  Methods& LPIPS (temp.)$\downarrow$ & WE(RAFT)$\downarrow$ & WE(alltracker)$\downarrow$ \\ 
  \midrule
  RelightVid~\cite{fang2025relightvid} & .02763& .00107& .00112\\
  UniRelight~\cite{he2025unirelight} & .02660& .00079& .00082\\
  DiffusionRenderer~\cite{DiffusionRenderer} & \cellcolor{yellow!15}.02629& \cellcolor{yellow!15}.00078& \cellcolor{yellow!15}.00080\\
  \midrule
  Ours & \cellcolor{red!15}.02532& \cellcolor{red!15}.00065& \cellcolor{red!15}.00060\\ \bottomrule
\end{tabular}
}
\par
\footnotesize
\colorbox{red!15}{\rule{0pt}{1em}\hspace{1em}} Best result \hspace{1em} \colorbox{yellow!15}{\rule{0pt}{1em}\hspace{1em}} Second-best result
\end{center}
\vspace{-2mm}
\end{table}

\subsection{Ablation Studies}
We conduct three ablation studies to investigate the effectiveness of key components in our proposed framework. All quantitative evaluations are performed on a representative subset of 200 test sequences randomly sampled from our \textit{Dynamic-subset}. Specifically, these sequences consist of paired flat-lit inputs and ground-truth OLAT videos, and all reported results (both quantitative and qualitative) evaluate the quality of the synthesized OLAT videos. To ensure fair comparison while managing computational costs, all ablated versions are trained for 20,000 steps with a sampling ratio of $0.1$ for images and $0.9$ for videos. Consistent with the main experiments, since our model focuses exclusively on the foreground subject, all quantitative metrics are computed on background-masked images to prevent background noise from biasing the results.

\begin{figure}
    \centering
    \includegraphics[width=\linewidth]{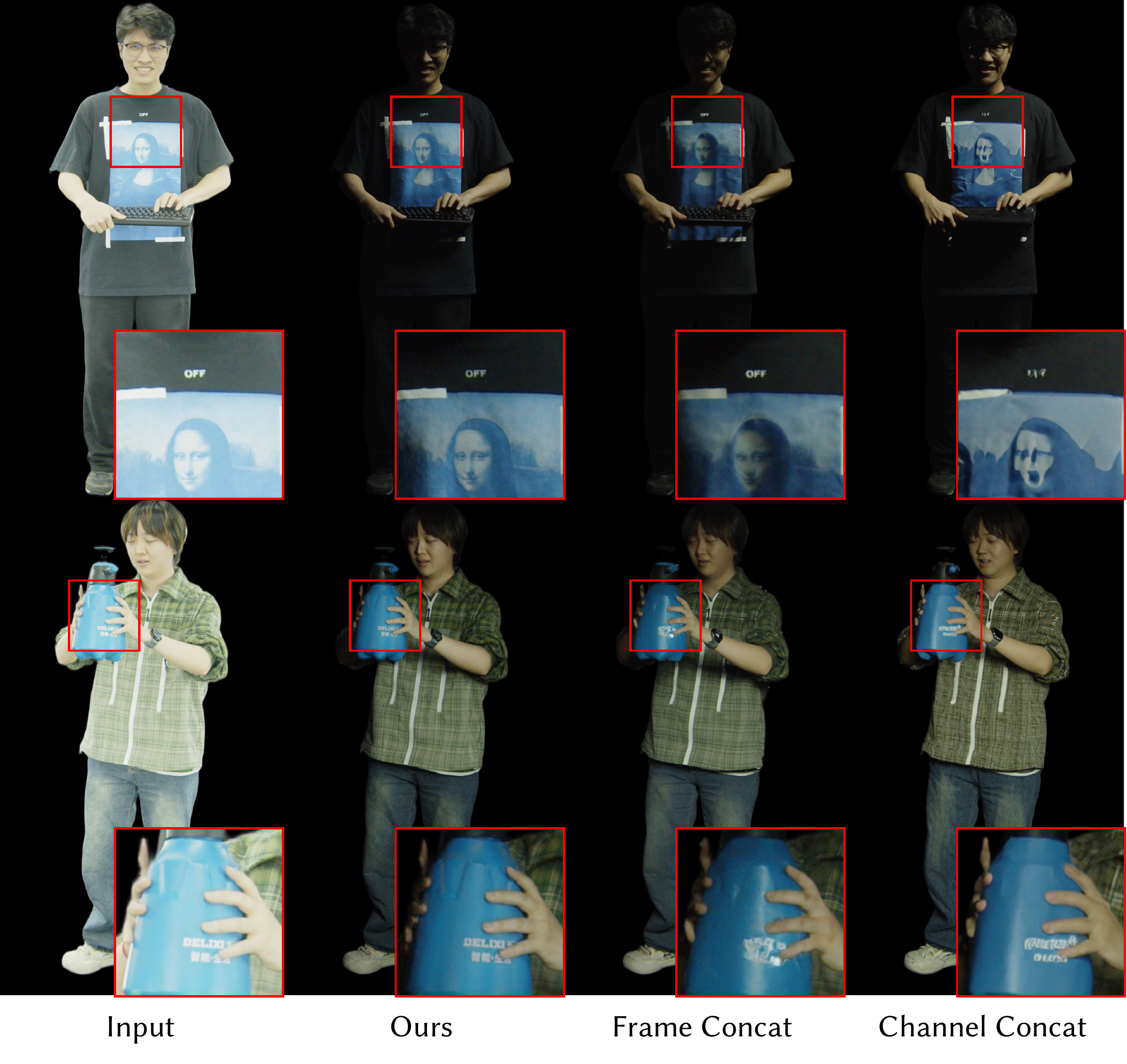}
    \vspace{-8mm}
    \caption{\textbf{Ablation study on input frames processing strategies.} We compare our method against concatenation along the channel and frame dimensions. Results show that Latent Bridge Matching strategy excels at preserving input details.}
    \Description{Comparison of channel concatenation, frame concatenation, and Latent Bridge Matching, with enlarged regions showing differences in retained details.}
    \label{fig:ablation_frame}
    \vspace{-4mm}
\end{figure}

\paragraph{Latent Bridge Matching for Input Frames}
To validate the efficacy of the LBM paradigm~\cite{chadebec2025lbm}, we compare it against standard diffusion conditioning strategies that rely on noise concatenation.
Specifically, we implemented two baseline variants where the flat-lit input latents are concatenated with the noisy latents along either the channel dimension or the frame dimension.
For the channel-wise concatenation baseline, we expanded the input channels of the first convolutional layer by a factor of two and initialized the weights of the newly added channels to zero for training stability.
Quantitative results in Tab.~\ref{tab:ablation} and qualitative comparisons in Fig.~\ref{fig:ablation_frame} reveal a significant performance gap.
The noise concatenation methods tend to degrade source fidelity, often producing artifacts or losing fine-grained texture details due to the destructive nature of the initial Gaussian noise.
In contrast, by establishing a direct transport trajectory from the source, bridge matching for video latents effectively preserves the structural integrity and high-frequency details of the input, yielding significantly higher fidelity in the final relit video.

\begin{table}[h]
\caption{\textbf{Quantitative comparison of ablated variants.}
We evaluate OLAT prediction under alternative input-frame and lighting-conditioning designs. The full model achieves the best overall performance across the reported perceptual and structural metrics.}
\label{tab:ablation}
\begin{center}
\small
\begin{tabular}{l c c c c }
\toprule
  Methods& PSNR$\uparrow$ & SSIM$\uparrow$ & LPIPS$\downarrow$ & FLIP$\downarrow$ \\\midrule 
  Cross Attn& 24.94& .9364& .0981&.0696\\
  Global AdaLN& 26.80& .9473& .0780&.0596\\
  \midrule
  Channel Concat& 27.13& .9434& .0815& .0589\\
  Frame Concat& \cellcolor{red!15}27.67& \cellcolor{yellow!15}.9481 & \cellcolor{yellow!15}.0750 & \cellcolor{yellow!15}.0556\\
  \midrule
  Ours & \cellcolor{yellow!15}27.61& \cellcolor{red!15}.9514& \cellcolor{red!15}.0727& \cellcolor{red!15}.0554\\ \bottomrule
\end{tabular}
\par
\footnotesize
\colorbox{red!15}{\rule{0pt}{1em}\hspace{1em}} Best result \hspace{1em} \colorbox{yellow!15}{\rule{0pt}{1em}\hspace{1em}} Second-best result
\end{center}
\vspace{-2mm}
\end{table}

\begin{figure}[t]
    \centering
    \includegraphics[width=\linewidth]{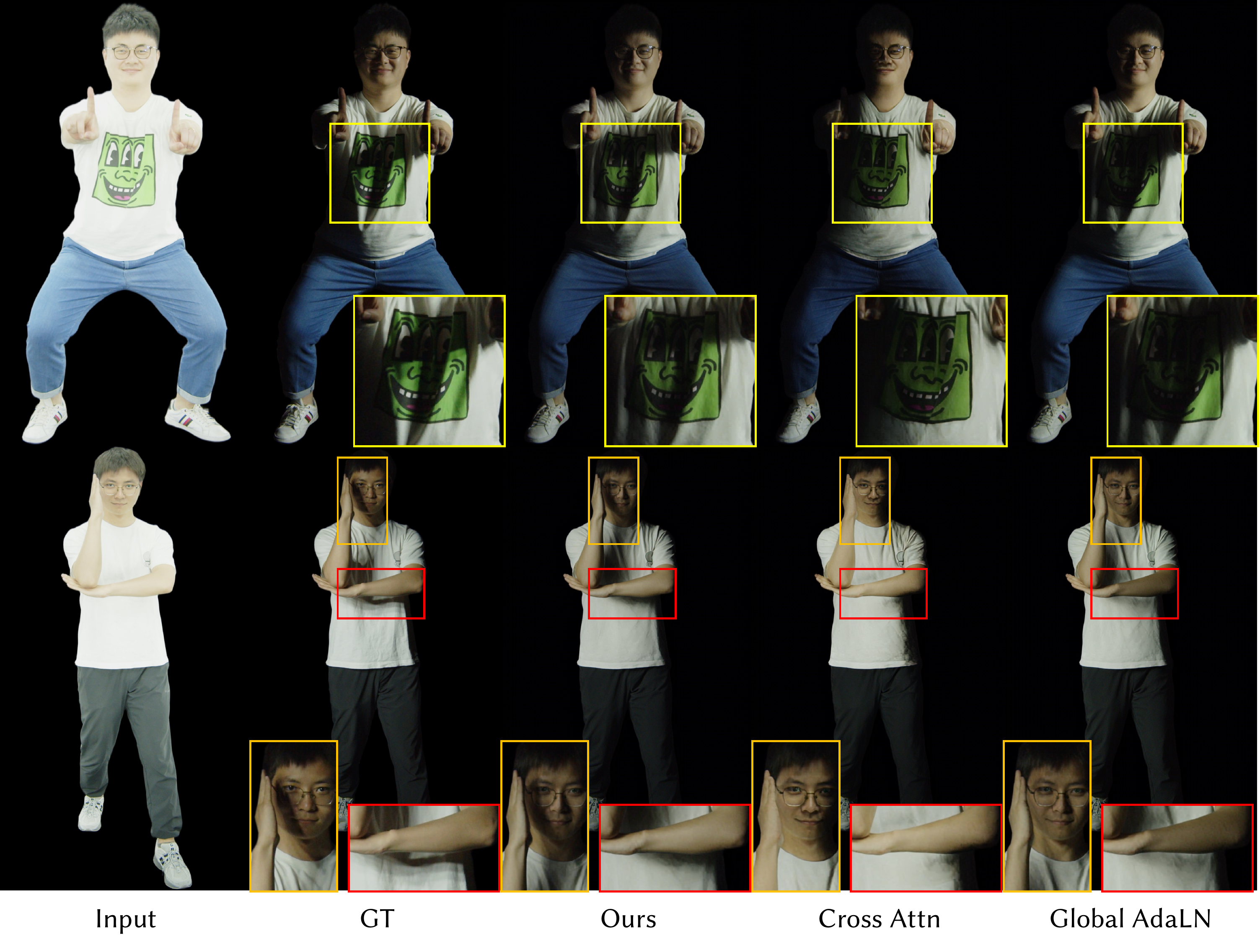}
    \vspace{-8mm}
    \caption{\textbf{Ablation study on lighting conditioning mechanisms.} We compare cross-attention, global AdaLN, and the proposed block-specific AdaLN strategy. Our result better matches the target OLAT image in the shown shadow placement and local details.}
    \Description{Lighting-conditioning comparison among cross-attention, global AdaLN, and block-specific AdaLN against a target OLAT image.}
    \label{fig:ablation_light}
    \vspace{-4mm}
\end{figure}

\begin{figure}
    \flushleft
    \includegraphics[width=0.92\linewidth]{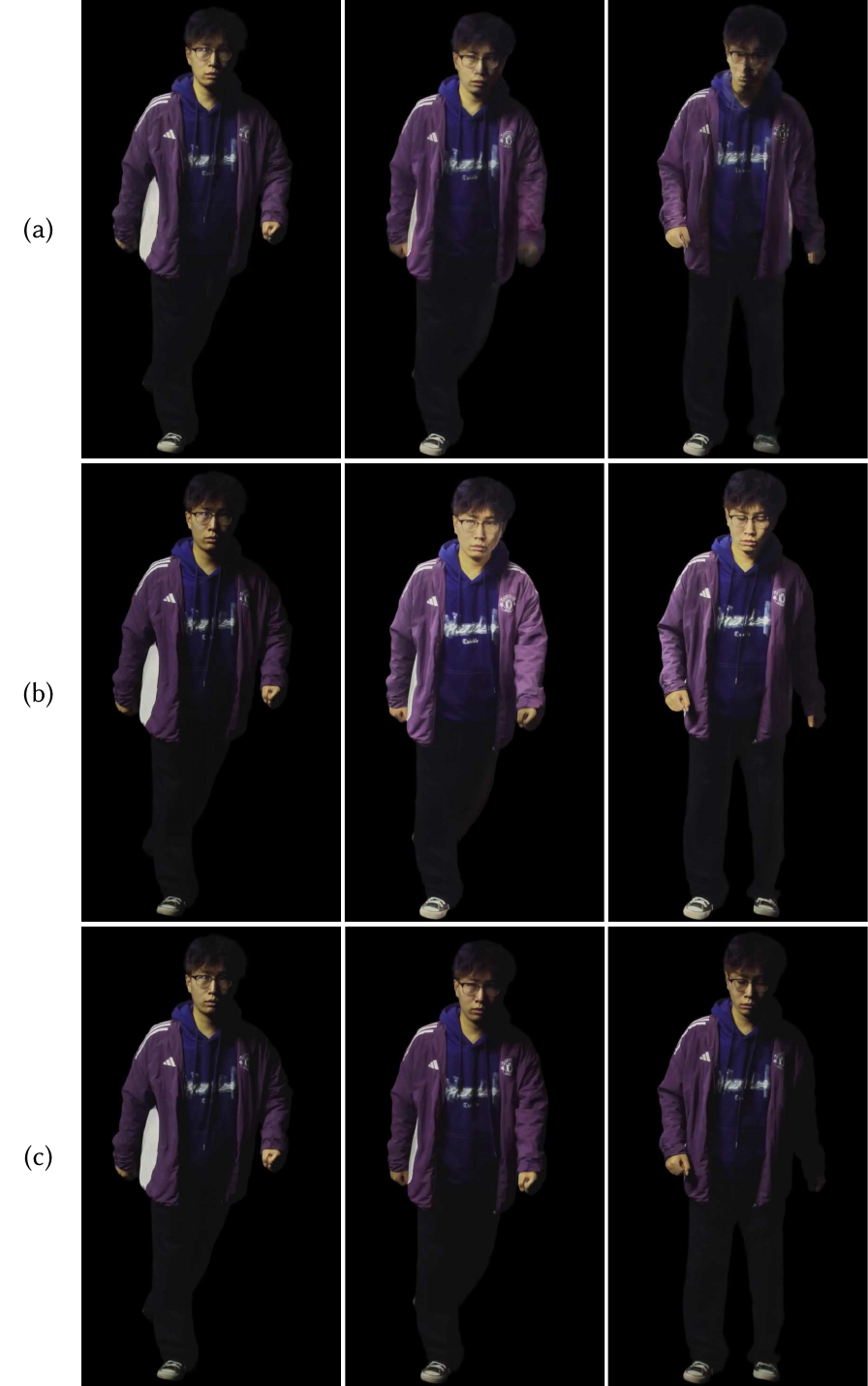}
    \caption{\textbf{Ablation study on strategies for length extrapolation:} (a) position interpolation~\cite{chen2023extendingcontextwindowlarge}, (b) sliding iterative sampling~\cite{jin2025diffuman4d}, and (c) our recurrent bridge context. Results show that our method avoids the visible artifacts and lighting drift of the two alternatives.}
    \Description{Frames from long-video inference using position interpolation, sliding iterative sampling, and recurrent bridge context, showing artifacts and lighting drift in the first two methods.}
    \label{fig:ablation_longvid}
    \vspace{-4mm}
\end{figure}

\paragraph{Lighting Conditioning}
We primarily evaluate our design against two alternative mechanisms: cross-attention and global AdaLN.
The cross-attention method, following~\citet{he2024diffrelight}, injects SH encodings by replacing the original text prompt embeddings, which risks disrupting the inherent input distribution of the backbone.
The global AdaLN scheme employs a single shared projection layer for all blocks, offering only coarse, uniform control.
In contrast, our block-specific AdaLN strategy preserves the integrity of the input features while enabling finer-grained, block-wise lighting modulation.
Quantitative results in Tab.~\ref{tab:ablation} and qualitative comparisons in Fig.~\ref{fig:ablation_light} demonstrate that our proposed conditioning scheme achieves more precise directional control and yields more photorealistic predictions.

\paragraph{Long Sequence Inference}
We compare our recurrent bridge context with Position Interpolation (PI)~\cite{chen2023extendingcontextwindowlarge} and sliding iterative sampling~\cite{jin2025diffuman4d} in Fig.~\ref{fig:ablation_longvid}.
PI introduces out-of-distribution positional embeddings, while heterogeneous latent timesteps cause significant lighting drift in sliding iterative sampling.

\subsection{Additional Qualitative Results}
\label{subsec:more_results}

\begin{figure*}
    \centering
    \includegraphics[width=\linewidth]{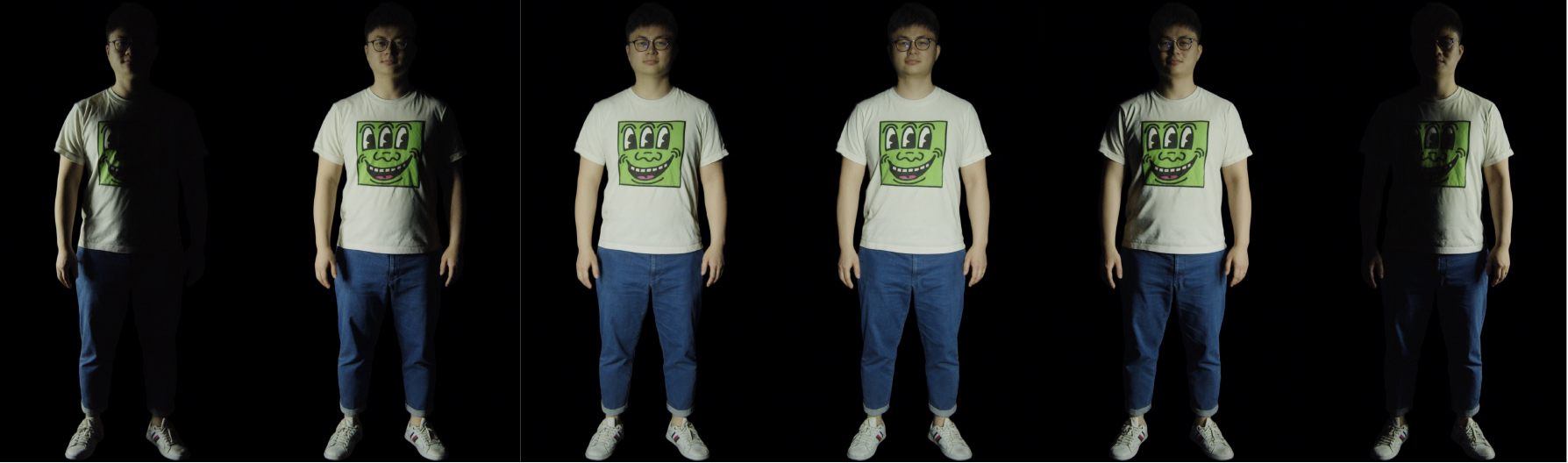}
    \vspace{-4mm}
    \caption{\textbf{OLAT synthesis under rotating lighting directions.} We visualize the predicted OLAT results for a specific video frame of an unseen subject under varying lighting directions. Note the physically plausible movement of cast shadows and the preservation of high-frequency clothing details as the light source rotates.}
    \Description{One subject relit by a sequence of rotating target-light directions, with corresponding changes in cast-shadow position.}
    \label{fig:olat_rotate}
\end{figure*}

\begin{figure*}
    \centering
    \includegraphics[width=\linewidth]{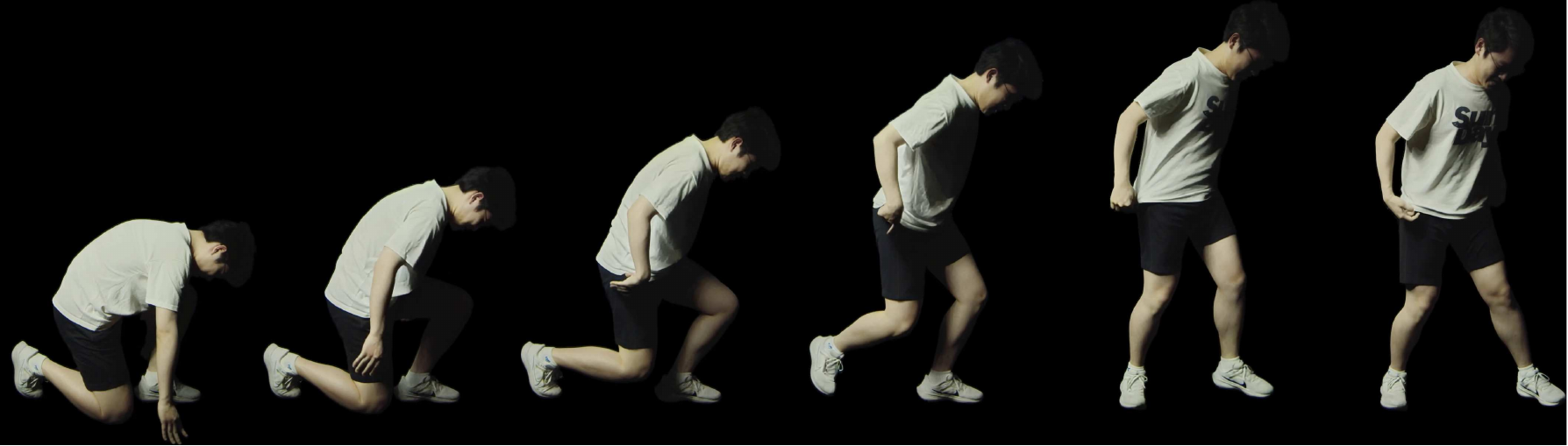}
    \vspace{-4mm}
    \caption{\textbf{Temporal consistency in dynamic OLAT generation.} Selected frames from a generated OLAT video sequence under a fixed directional light, sampled at 10-frame intervals. Our model maintains temporally coherent shading throughout the complex motion.}
    \Description{Frames sampled at 10-frame intervals from a dynamic OLAT video under a fixed directional light.}
    \label{fig:olat_motion}
\end{figure*}

\begin{figure*}
    \centering
    \includegraphics[width=\linewidth]{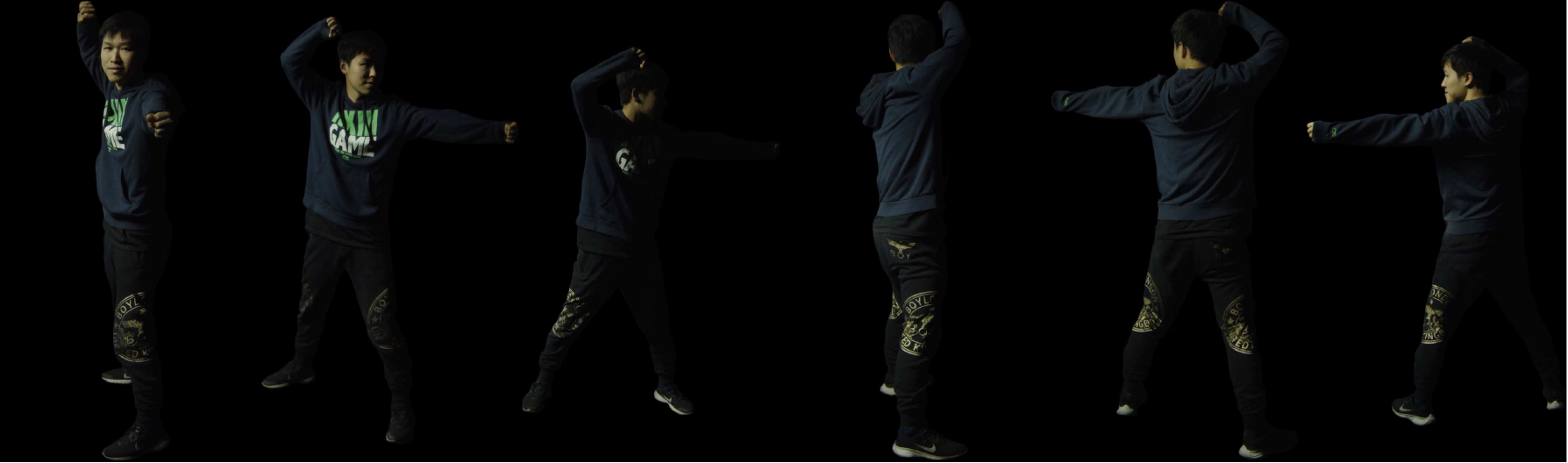}
    \vspace{-4mm}
    \caption{\textbf{Generalization to arbitrary viewpoints.} We present OLAT predictions for the same subject captured from distinct camera views. For each view, the model successfully synthesizes the corresponding OLAT image under a specified lighting direction, demonstrating robust generalization across diverse observation viewpoints.}
    \Description{Predicted OLAT images of the same subject from several captured camera viewpoints under one specified light direction.}
    \label{fig:olat_view}
\end{figure*}

We further demonstrate the versatility and robustness of our model on unseen subjects. 
As illustrated in Fig.~\ref{fig:olat_rotate}, our model synthesizes precise shading and self-shadowing effects under continuously rotating lighting directions, maintaining physical plausibility.
Fig.~\ref{fig:olat_motion} validates temporal stability, showcasing a dynamic performance coherently illuminated by a fixed directional light source across the sequence.
Finally, Fig.~\ref{fig:olat_view} highlights the geometric consistency of our predictions, successfully generalizing to diverse camera viewpoints for a specified lighting direction.
For more extensive visualization results and dynamic comparisons, please refer to our supplementary video.

\section{Conclusion}
\label{sec:conclusion}
In this work, we have presented a comprehensive framework for controllable dynamic full-body relighting from monocular flat-lit videos.
By shifting the paradigm from direct relit-image generation to data-driven OLAT video prediction, our approach decomposes human light transport into directional basis responses and reconstructs the final appearance through linear-space intensity weighting.
This formulation makes direction explicit in the OLAT bases and relative intensity explicit in the environment-map weights while preserving appearance details and temporal coherence.
The accompanying large-scale dynamic full-body OLAT dataset provides the directional supervision needed to learn these responses across diverse identities, outfits, motions, and views.

Current limitations include the flat-lit input assumption, sensitivity of delighting to strong shadows or colored illumination, and the reconstruction ceiling of the pretrained video VAE.
The explicit OLAT representation also incurs basis-dependent generation and storage costs, while cross-OLAT and cross-view consistency remains implicit rather than architecturally guaranteed.
Finally, the finite-area capture lights limit directional resolution; missing dynamic relighting ground truth and reliable pixel-level lighting-angle annotations further constrain quantitative evaluation.
Further discussion and failure analysis are provided in the supplementary material.

These boundaries motivate more robust delighting, higher-fidelity video autoencoders, efficient basis prediction with explicit consistency, finer illumination capture, and dynamic lighting benchmarks.
The dataset may further support neural rendering and controllable generative video editing.

\begin{acks}
We thank Dongli Tan for LightStage camera calibration and Linxian Wu for providing paired outdoor environment panoramas and reference images captured under the same lighting conditions.

This work was partially supported by National Key R\&D Program of China (No.~2024YFB2809105), NSFC (No.~U24B20154), Zhejiang Provincial Natural Science Foundation of China (No.~LR25F020003), and Information Technology Center and State Key Lab of CAD\&CG, Zhejiang University.
\end{acks}

\clearpage
\bibliographystyle{ACM-Reference-Format}
\bibliography{references}

\clearpage
\appendix
\twocolumn[{%
    \centering
    \Large \textbf{Supplementary Material} \\[1em]
}]
\section{Dataset Statistics}
\label{sec:supp_dataset_statistics}

Participants were 18--30 years old. Tab.~\ref{tab:dataset-statistics} summarizes the combined dataset composition.

\definecolor{datasetstatbar}{RGB}{74, 111, 165}
\definecolor{datasetstatdetailbar}{RGB}{137, 164, 201}
\newcommand{\datasetstatbar}[1]{\makebox[14mm][l]{\textcolor{datasetstatbar}{\rule{#1}{1.05ex}}}}
\newcommand{\datasetstatdetailbar}[1]{\makebox[14mm][l]{\textcolor{datasetstatdetailbar}{\rule{#1}{1.05ex}}}}

\newcommand{\DatasetStatsGenderRows}{%
Male & \textbf{135} & \datasetstatbar{8.44mm} \\
Female & \textbf{89} & \datasetstatbar{5.56mm} \\
}

\newcommand{\DatasetStatsAccessoryRows}{%
None & \textbf{118} & \datasetstatbar{7.38mm} \\
Any accessory/prop & \textbf{106} & \datasetstatbar{6.62mm} \\
}

\newcommand{\DatasetStatsAccessoryDetailRows}{%
\quad Glasses & \textbf{87} & \datasetstatdetailbar{5.44mm} \\
\quad Handheld prop & \textbf{20} & \datasetstatdetailbar{1.25mm} \\
\quad Hat/cap & \textbf{8} & \datasetstatdetailbar{0.60mm} \\
\quad Other accessory/prop & \textbf{8} & \datasetstatdetailbar{0.60mm} \\
}

\newcommand{\DatasetStatsTopRows}{%
Outerwear & \textbf{93} & \datasetstatbar{5.81mm} \\
Long-sleeve & \textbf{84} & \datasetstatbar{5.25mm} \\
T-shirt & \textbf{38} & \datasetstatbar{2.38mm} \\
Dress/one-piece & \textbf{5} & \datasetstatbar{0.60mm} \\
Sleeveless & \textbf{3} & \datasetstatbar{0.60mm} \\
Other & \textbf{1} & \datasetstatbar{0.60mm} \\
}

\newcommand{\DatasetStatsBottomRows}{%
Full-length pants & \textbf{208} & \datasetstatbar{13.00mm} \\
Shorts & \textbf{11} & \datasetstatbar{0.69mm} \\
Skirt/dress & \textbf{5} & \datasetstatbar{0.60mm} \\
}

\newcommand{\DatasetStatsColorRows}{%
White/light & \textbf{45} & \datasetstatbar{2.81mm} \\
Gray & \textbf{40} & \datasetstatbar{2.50mm} \\
Multicolor & \textbf{39} & \datasetstatbar{2.44mm} \\
Black/dark & \textbf{33} & \datasetstatbar{2.06mm} \\
Blue & \textbf{23} & \datasetstatbar{1.44mm} \\
Yellow/brown & \textbf{22} & \datasetstatbar{1.38mm} \\
Green & \textbf{11} & \datasetstatbar{0.69mm} \\
Red/pink & \textbf{7} & \datasetstatbar{0.60mm} \\
Purple & \textbf{4} & \datasetstatbar{0.60mm} \\
}

\begin{table*}[t]
    \caption{\textbf{Combined dataset composition.} All counts refer to the 224 identity-outfit combinations, not to unique participants. For accessories and props, \textit{None} and \textit{Any accessory/prop} are mutually exclusive; indented rows are multi-label breakdowns of \textit{Any accessory/prop} and may overlap.}
    \label{tab:dataset-statistics}
    \centering
    \begingroup
    \small
    \setlength{\tabcolsep}{2pt}
    \renewcommand{\arraystretch}{1.08}
    \begin{minipage}[t]{0.315\textwidth}
        \vspace{0pt}
        \begin{tabular*}{\linewidth}{@{\extracolsep{\fill}} l r l @{}}
            \toprule
            Category & Count & Distribution \\
            \midrule
            \multicolumn{3}{@{}l}{\textbf{Gender}} \\
            \DatasetStatsGenderRows
            \addlinespace[2pt]
            \multicolumn{3}{@{}l}{\textbf{Accessory / prop}} \\
            \DatasetStatsAccessoryRows
            \DatasetStatsAccessoryDetailRows
            \multicolumn{3}{@{}l}{\phantom{\textbf{Bottom garment}}} \\
            \bottomrule
        \end{tabular*}
    \end{minipage}\hfill
    \begin{minipage}[t]{0.315\textwidth}
        \vspace{0pt}
        \begin{tabular*}{\linewidth}{@{\extracolsep{\fill}} l r l @{}}
            \toprule
            Category & Count & Distribution \\
            \midrule
            \multicolumn{3}{@{}l}{\textbf{Top garment}} \\
            \DatasetStatsTopRows
            \addlinespace[2pt]
            \multicolumn{3}{@{}l}{\textbf{Bottom garment}} \\
            \DatasetStatsBottomRows
            \bottomrule
        \end{tabular*}
    \end{minipage}\hfill
    \begin{minipage}[t]{0.315\textwidth}
        \vspace{0pt}
        \begin{tabular*}{\linewidth}{@{\extracolsep{\fill}} l r l @{}}
            \toprule
            Category & Count & Distribution \\
            \midrule
            \multicolumn{3}{@{}l}{\textbf{Dominant top color}} \\
            \DatasetStatsColorRows
            \multicolumn{3}{@{}l}{\phantom{\textbf{Bottom garment}}} \\
            \bottomrule
        \end{tabular*}
    \end{minipage}
    \endgroup
\end{table*}

\section{Supplementary of Data Processing}
\subsection{Rolling Shutter Detection and Sequence Segmentation}
\begin{figure}
    \centering
    \includegraphics[width=\linewidth]{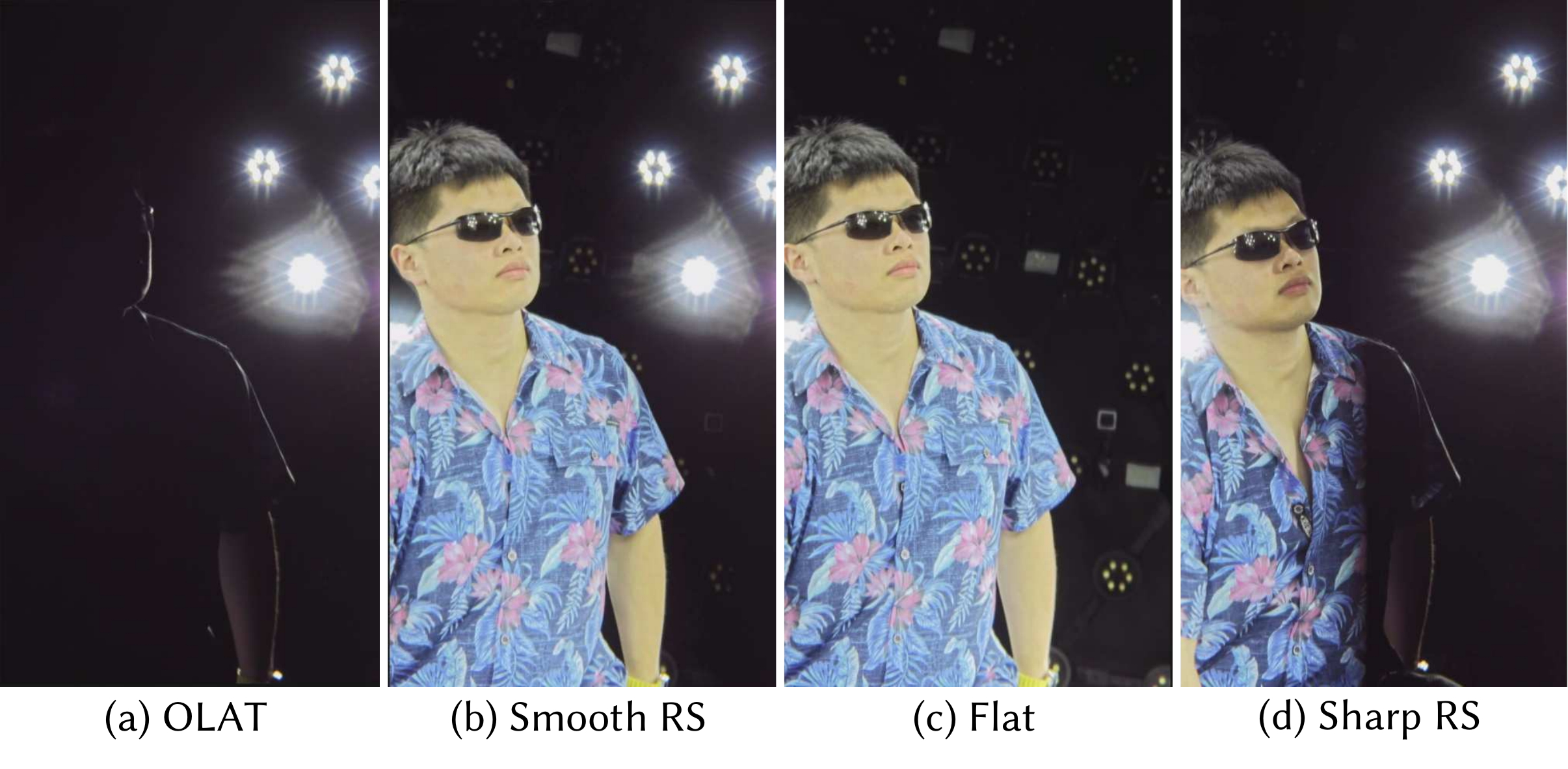}
    \caption{\textbf{Analysis of rolling shutter artifacts.}
    {(a--c)} Three consecutive frames contain an OLAT frame, a rolling-shutter (RS) frame, and a flat-lit frame.
    {(d)} An example of a "distinct" RS frame from the same sequence where the exposure boundary is clearly visible.
    Due to the vertical orientation of our sensors, the electronic rolling shutter scans horizontally (right-to-left). Ideally, this results in a sharp transition boundary as seen in (d). 
    However, as shown in {(b)}, high-frequency boundary lines are often smoothed out by video compression algorithms.
    Additionally, {(a)} illustrates a failure case for heuristic brightness-based detectors: direct opposing illumination causes pixel intensities to exceed those of the flat-lit frame, complicating simple thresholding methods.
    }
    \Description{Four frames illustrating rolling-shutter detection: an OLAT frame, an artifact-contaminated frame, a flat-lit frame, and a frame with a distinct exposure boundary.}
    \label{fig:dataset_rs_example}
\end{figure}

The light-switch cycle is intentionally set slightly longer than two camera frame periods (approximately 33.3 ms at 60 fps), which facilitates rolling-shutter detection.
In addition to the target flat-lit and OLAT frames, our raw footage inevitably contains frames affected by rolling-shutter (RS) artifacts.
Both subsets rely on the precise identification of valid frames to correctly extract lighting sequences. 
However, identifying RS frames is non-trivial for the \textit{Dynamic-subset}: due to H.264 compression, the distinct RS boundary is blurred by spatial artifacts (Fig.~\ref{fig:dataset_rs_example}(b)), rendering traditional spatial stitching or gradient-based detection infeasible.

\begin{figure}
    \centering
    \includegraphics[width=\linewidth]{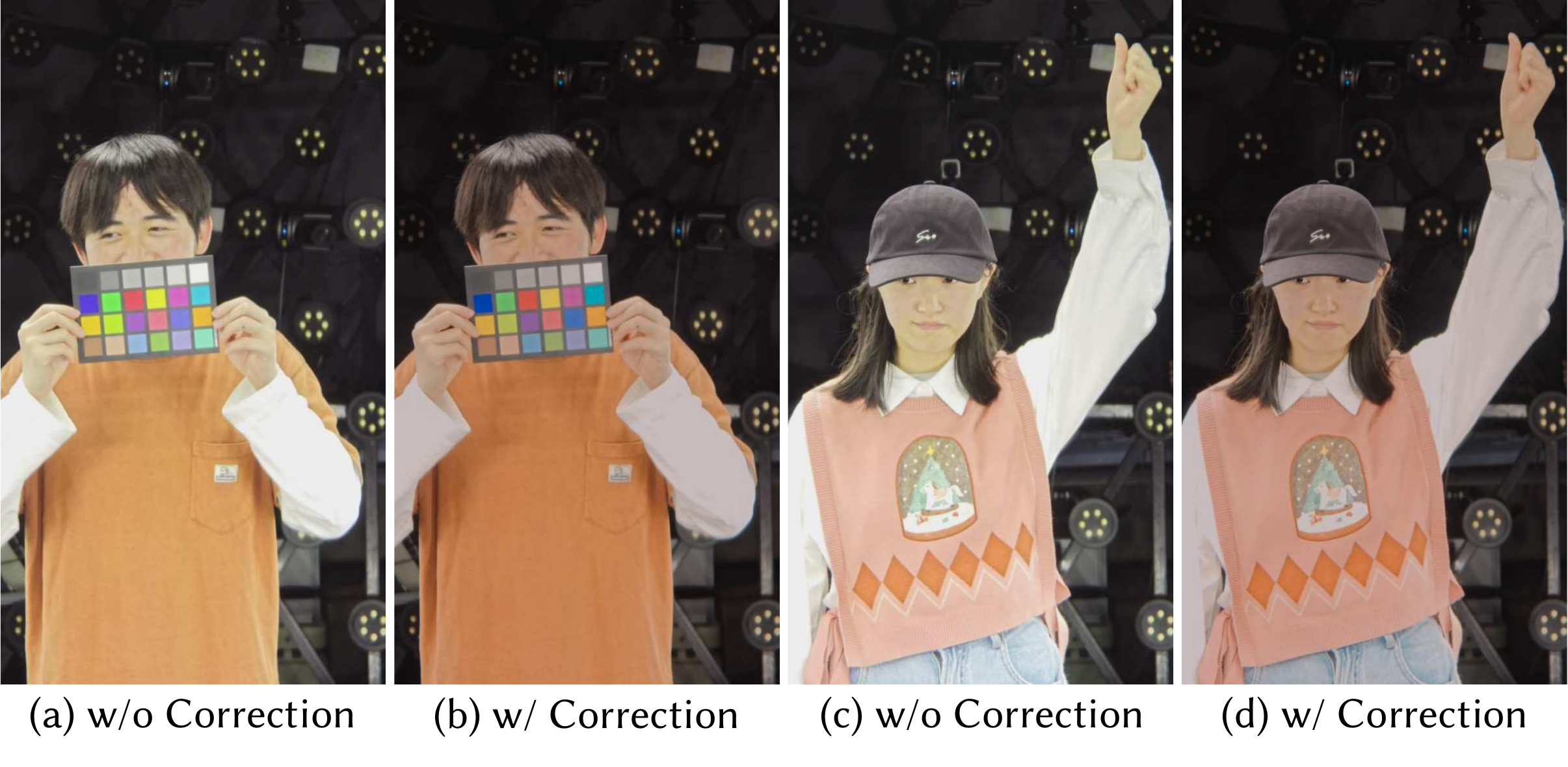}
    \caption{
    \textbf{Evaluation of Color Correction.} 
    (a) and (b) show the raw and color-corrected frames of the calibration sequence, respectively.
    (c) and (d) demonstrate the generalization of the computed CCM to a human subject within the same view.
    A significant artifact is observed in the white sleeve region in (c) and (d). 
    The CCM causes originally neutral white pixels (near saturation) to shift significantly towards a darker, tinted hue, violating the scene's white balance. 
    This severe deviation on neutral tones motivates our decision to keep CCMs optional.
    }
    \Description{Calibration and human-subject frames before and after color correction, with the corrected white sleeve exhibiting a dark color cast.}
    \label{fig:ColorCorrections}
\end{figure}

To distinguish valid frames from artifacts, we developed a deep learning-based classifier. 
We annotated 1,000 frames using three classes: flat-lit, OLAT, and rolling shutter.
An EfficientNet-based~\cite{tan2020efficientnetrethinkingmodelscaling} classifier was trained with label smoothing to mitigate the small sample size. 
Crucially, we modified the network to accept high-resolution inputs ($512\times512$) to preserve the high-frequency features of the RS bands, avoiding aliasing during downsampling.
This robust classifier achieves near-perfect accuracy, serving as the foundation for our sequence segmentation:
For the \textit{Dynamic-subset}, leveraging the frame-wise classifications, we segment the video stream into 51 discrete sequences, strictly adhering to the fixed structure of 18 interleaved flat-lit--OLAT pairs per sequence.
For the \textit{Static-subset}, the classifier identifies the extended intervals of flat-lit frames (approx. 120 frames) that act as delimiters between distinct static poses.

\subsection{Optional Color Calibration.}
Although we computed Color Correction Matrices (CCMs) using X-Rite charts, applying the full CCMs can introduce severe color casts, especially by shifting white regions toward darker hues (see Fig.~\ref{fig:ColorCorrections}).
We therefore provide the CCMs as optional metadata rather than applying them to the primary training data.

\section{Inference-time Processing}
\label{sec:supp_inference_processing}

\paragraph{OLAT post-processing.}
The dynamic range of generated videos differs from that of common Internet videos and may leave weak responses in OLAT bases that should be nearly dark.
Therefore, after OLAT prediction and before linear environment-map assembly, we apply a deterministic foreground-aware post-processing step to the predicted OLAT videos.
Within the foreground, we define brightness as the maximum RGB value and quadratically attenuate responses below $0.1$.
For directions prone to back-light leakage, we apply stronger sixth-power attenuation below $0.3$ together with a $5\times5$ Gaussian blur. The directional risk score is centered at an azimuth of $-18^\circ$, is maximal over $[-36^\circ,0^\circ]$, and decreases with a cubic smoothstep to zero at an angular distance of $36^\circ$; the zenith basis is excluded.
We blend the default quadratic and stronger corrections according to this risk score while leaving pixels outside the foreground mask unchanged.
This procedure reduces dark noise and spurious back-light leakage while preserving bright, valid light responses used for relighting.

\begin{figure}
    \centering
    \includegraphics[width=\linewidth]{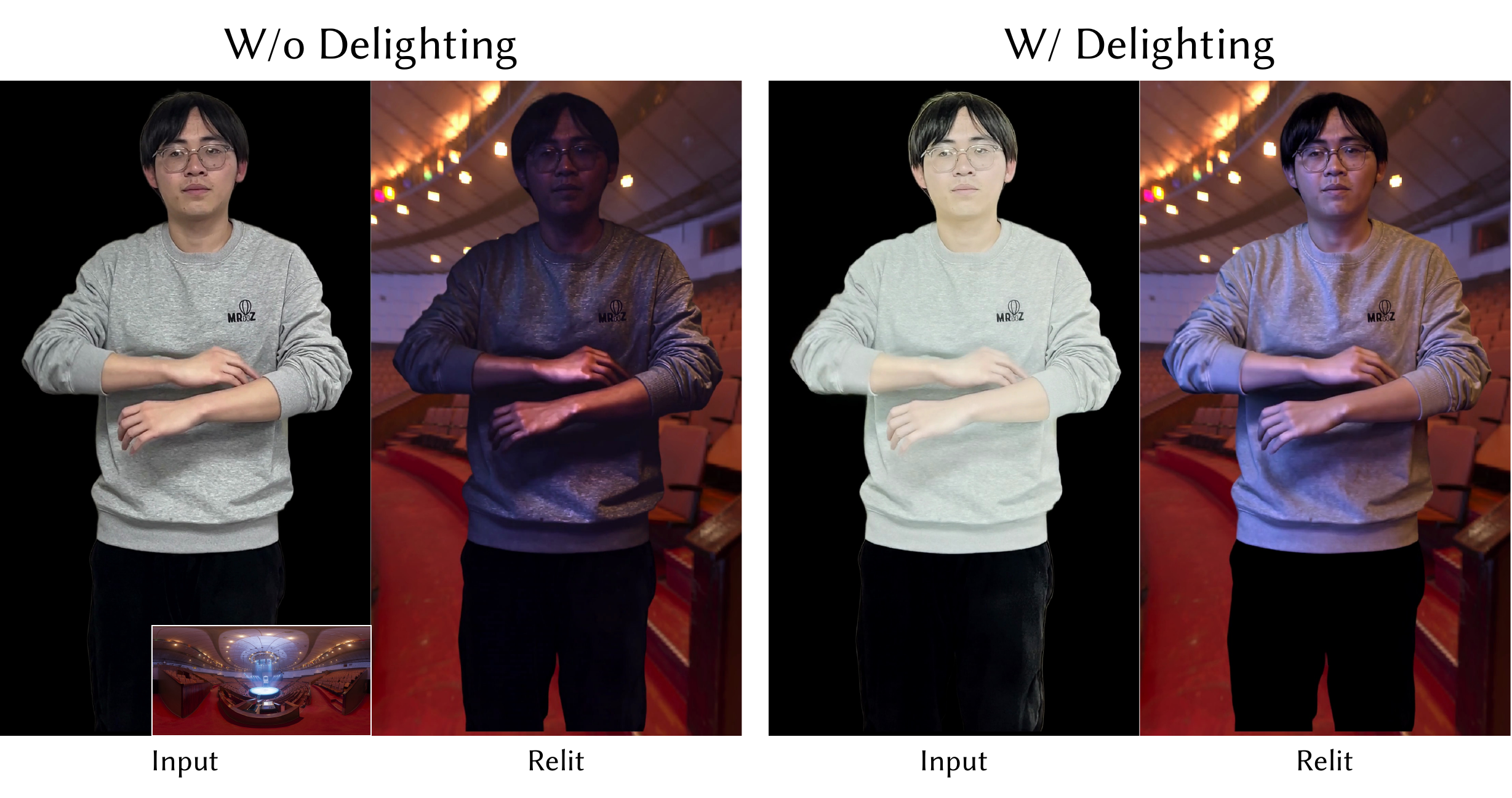}
    \caption{\textbf{Effect of the delighting module.} The comparison illustrates that the module adapts in-the-wild inputs toward the flat-lit domain used by our OLAT predictor, making downstream relighting more consistent with the training setting.}
    \Description{Comparison of an in-the-wild input, its delighted version, and downstream relighting results, showing reduced residual illumination after delighting.}
    \label{fig:delight_effect}
\end{figure}

\paragraph{Delighting for in-the-wild inputs.}
Since our main OLAT predictor is trained on controlled flat-lit videos, in-the-wild videos with cast shadows or colored illumination can introduce a domain gap.
For in-the-wild testing, we first pass the input video through a lightweight delighting module that maps uncontrolled illumination toward our flat-lit training domain.
The module is implemented as a LoRA adaptation of Wan2.1 1.3B~\cite{wan2025}, trained as a video-to-video translation model using our generated relit videos as inputs and the corresponding flat-lit videos as targets.
We use a rank-$32$ LoRA and a 16-step LBM schedule with bridge-noise standard deviation $0.001$.
Using AdamW with a constant learning-rate schedule, we train on more than 1,000 paired clips for 2,000 steps with a learning rate of $10^{-4}$ and a global batch size of $8$.
We train separate 57-frame video and single-frame variants under the same settings, and use four LBM steps at inference.
As shown in Fig.~\ref{fig:delight_effect}, this preprocessing suppresses pre-existing illumination patterns before OLAT prediction, improving robustness on real-world videos without changing the core relighting model.

\paragraph{Foreground compositing.}
For all visualizations that composite relit subjects onto target backgrounds, we refine the foreground masks using BiRefNet's \texttt{refine\_foreground} option~\cite{zheng2024birefnet} before alpha compositing.
This step improves high-resolution foreground boundaries and reduces dark contour artifacts around the subjects.

\section{Limitations and Failure Analysis}
\label{sec:vae_analysis}

\begin{figure}[!t]
    \centering
    \includegraphics[width=0.78\linewidth]{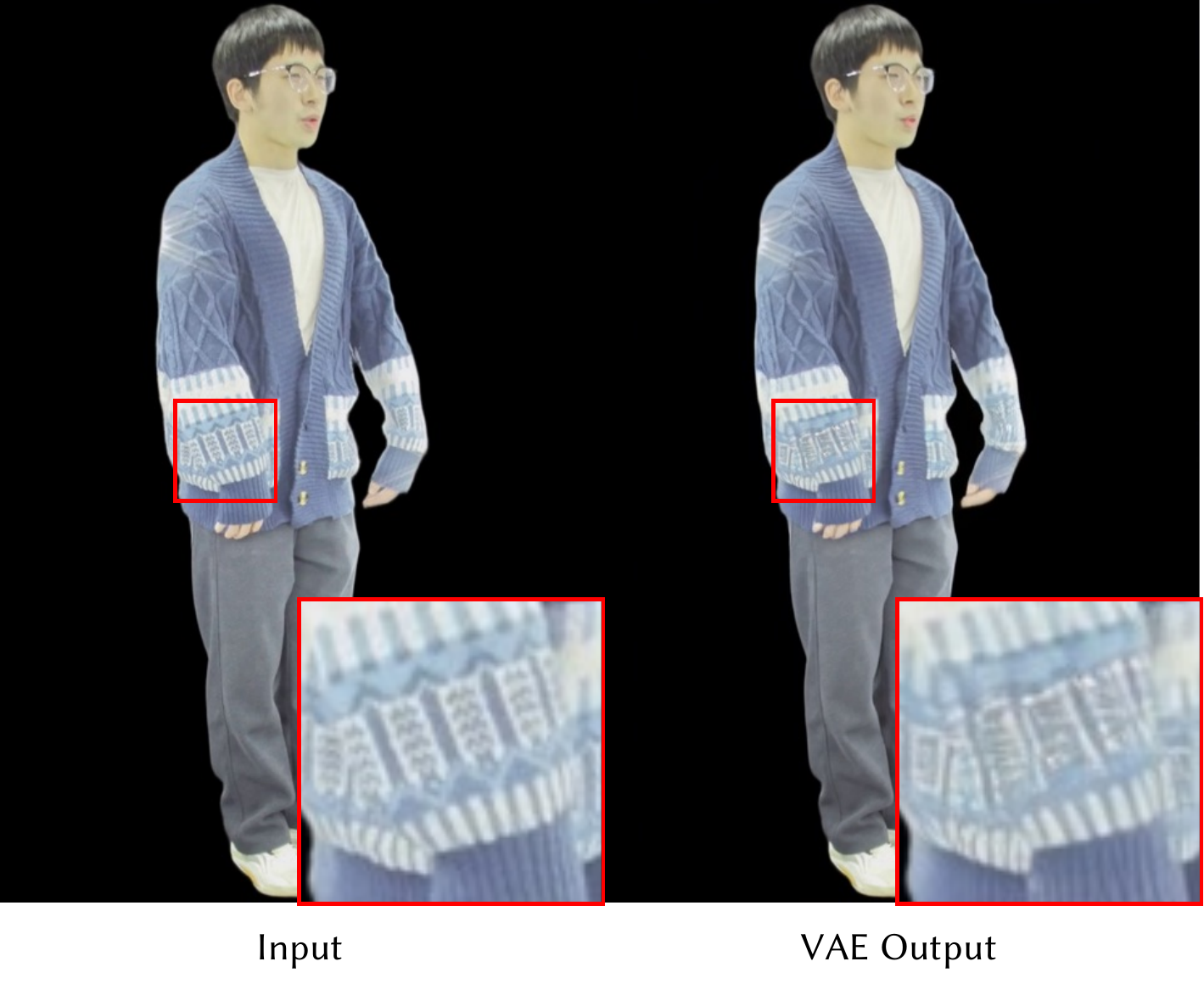}
    \caption{\textbf{Reconstruction limitations of the video VAE.} We compare the ground-truth input frame (left) with the VAE reconstruction (right). The reconstruction exhibits a noticeable loss of high-frequency details, particularly in regions with rapid motion.}
    \Description{Side-by-side comparison of an input frame and its video VAE reconstruction, with enlarged regions showing softened facial and fabric details.}
    \label{fig:vae}
\end{figure}

A key component of our framework is the latent video diffusion model, which relies on a pretrained video VAE for encoding and decoding frames.
Consequently, output fidelity is bounded by the reconstruction quality of this VAE.
We conducted an analysis to isolate this bottleneck. As illustrated in Fig.~\ref{fig:vae}, while the VAE performs adequately for static or slow-moving content, it exhibits a noticeable loss of high-frequency details in scenarios involving rapid or large-scale motions (e.g., blurring of facial features or fabric textures).
This degradation is inherent to the compression artifacts of the VAE and is independent of our relighting prediction. Future improvements in video autoencoder architectures will directly benefit our framework's output fidelity.

The flat-lit input assumption is practical for controlled production but limits direct use on arbitrary consumer footage.
Although the delighting module reduces this domain gap, it remains preliminary: cast shadows or strong colored illumination may not be fully removed and can be interpreted by the OLAT predictor as material- or geometry-dependent responses, propagating residual artifacts to the relit outputs.
The in-the-wild results therefore demonstrate feasibility under varied inputs rather than uniform robustness to uncontrolled illumination.

Controllability through explicit OLAT bases also introduces computational and consistency tradeoffs.
Generation time and storage grow approximately linearly with the number of lighting directions; parallel prediction reduces wall-clock time and cached bases can be reused, but neither removes the total prediction and storage burden.
Because bases and views are generated independently, their consistency arises implicitly from the common flat-lit input and training data rather than from an explicit architectural constraint.

The capture system imposes complementary representation and evaluation limits.
Each One-Light group spans a finite area, which smooths the incident lighting direction and may under-resolve sharp specular responses.
Moreover, the system cannot record dynamic sequences under arbitrary environment illumination, leaving no dynamic ground-truth relit videos for target-light evaluation. Accordingly, static image metrics measure image-level agreement under target lighting, whereas the reported video metrics measure temporal smoothness rather than target-light correctness.
Precise angular evaluation of directional control is also difficult because a single ground-truth lighting angle cannot be assigned reliably to the spatially varying shadows and highlights on a deforming human; we therefore do not report a shadow-direction offset metric.

\section{Additional Experimental Results}
\label{sec:supp_additional_results}

\begin{figure*}[t]
    \centering
    \begin{minipage}[t]{0.47\textwidth}
        \centering
        \includegraphics[width=\linewidth]{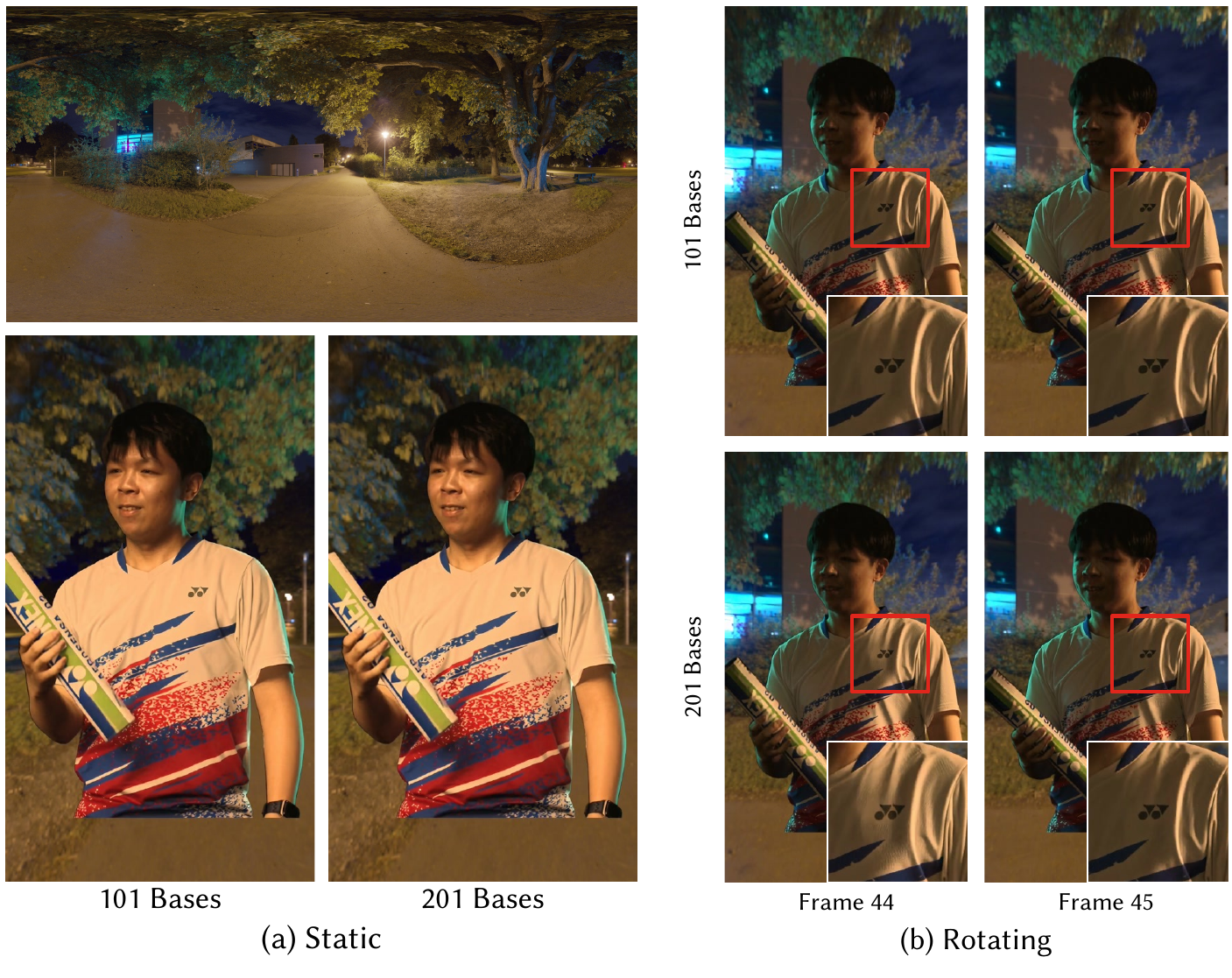}
        \captionof{figure}{\textbf{Quality--cost tradeoff between 101 and 201 OLAT bases.} (a) Under a static environment map, 101 and 201 bases provide comparable visual quality. (b) For a rotating environment map, 201 bases respond more smoothly to lighting changes; this is visible in the evolution of the clothing shadow between Frames 44 and 45.}
        \Description{Static and rotating-light comparisons of relighting with 101 and 201 OLAT bases, including consecutive frames that show the shadow transition.}
        \label{fig:supp_bases_comparison}
    \end{minipage}
    \hfill
    \begin{minipage}[t]{0.47\textwidth}
        \centering
        \includegraphics[width=\linewidth]{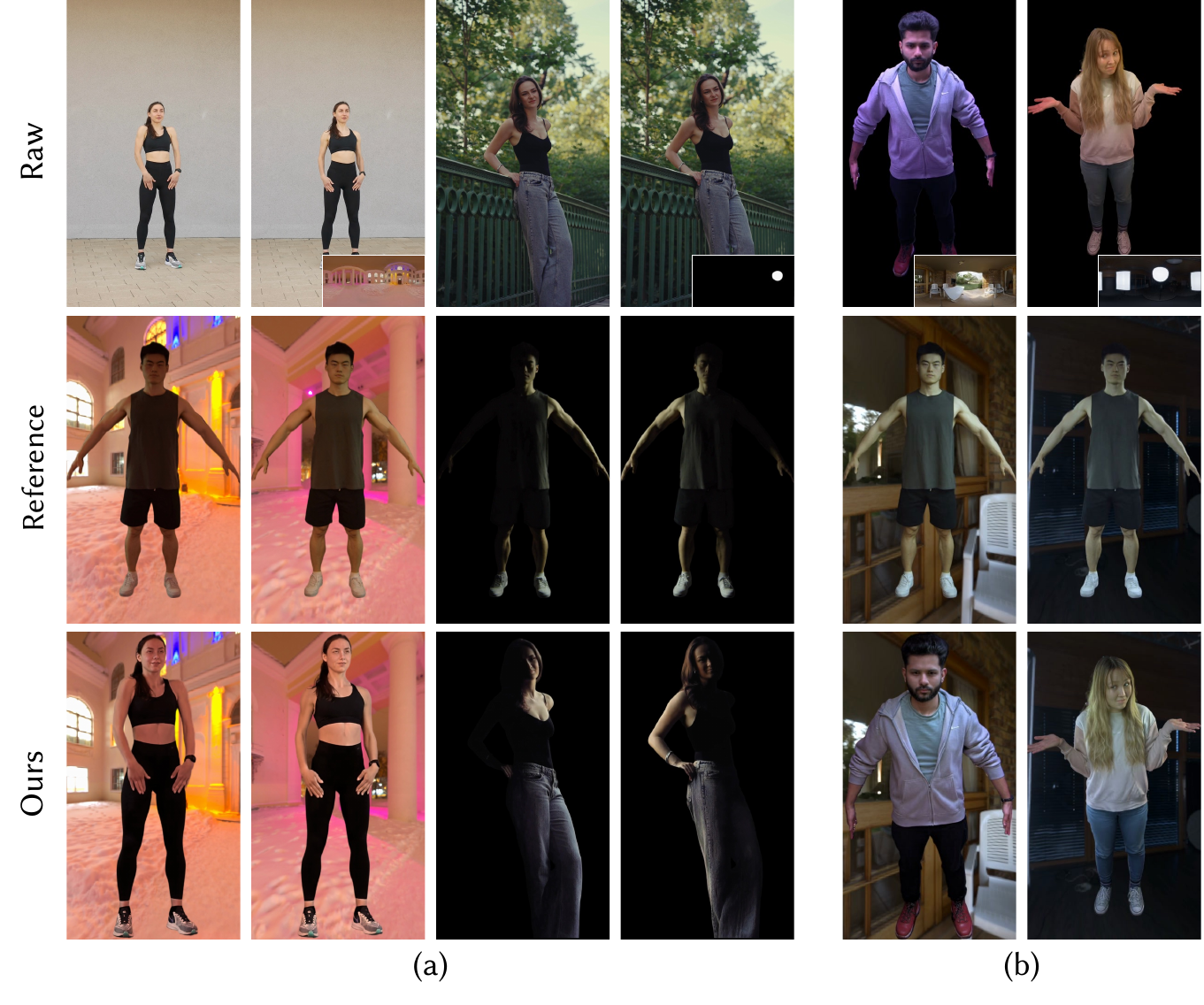}
        \captionof{figure}{\textbf{Additional in-the-wild relighting results.} (a) Video inputs sourced from Pexels~\cite{pexelsVideos} and HumanVid~\cite{wang2024humanvid}. (b) Images captured under environment illumination from HumanOLAT~\cite{teufelgera2025HumanOLAT}.}
        \Description{Rows of source inputs, target-light references, and relit outputs for examples from Pexels, HumanVid, and HumanOLAT.}
        \label{fig:supp_wild_results}
    \end{minipage}

    \vspace{3mm}

    \begin{minipage}{0.96\textwidth}
        \centering
        \includegraphics[width=\linewidth]{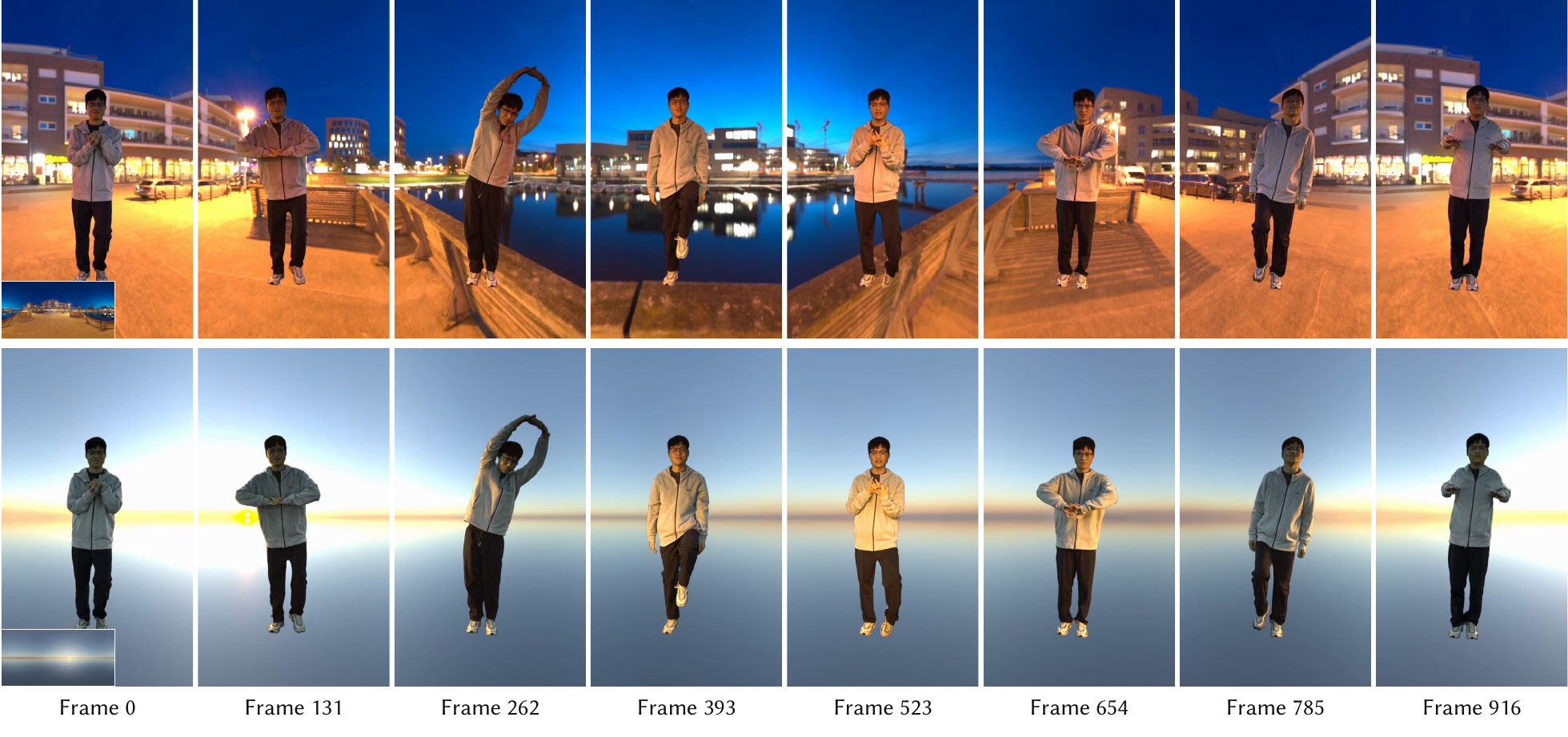}
        \captionof{figure}{\textbf{Long-sequence relighting.} We show two relit videos, each containing 917 frames, under two target environment maps. Both examples maintain stable subject appearance and lighting evolution over the complete sequences.}
        \Description{Sampled frames from two 917-frame relit sequences under different target environment maps.}
        \label{fig:supp_long_video}
    \end{minipage}
\end{figure*}

\begin{figure*}[t]
    \centering
    \includegraphics[width=0.98\textwidth]{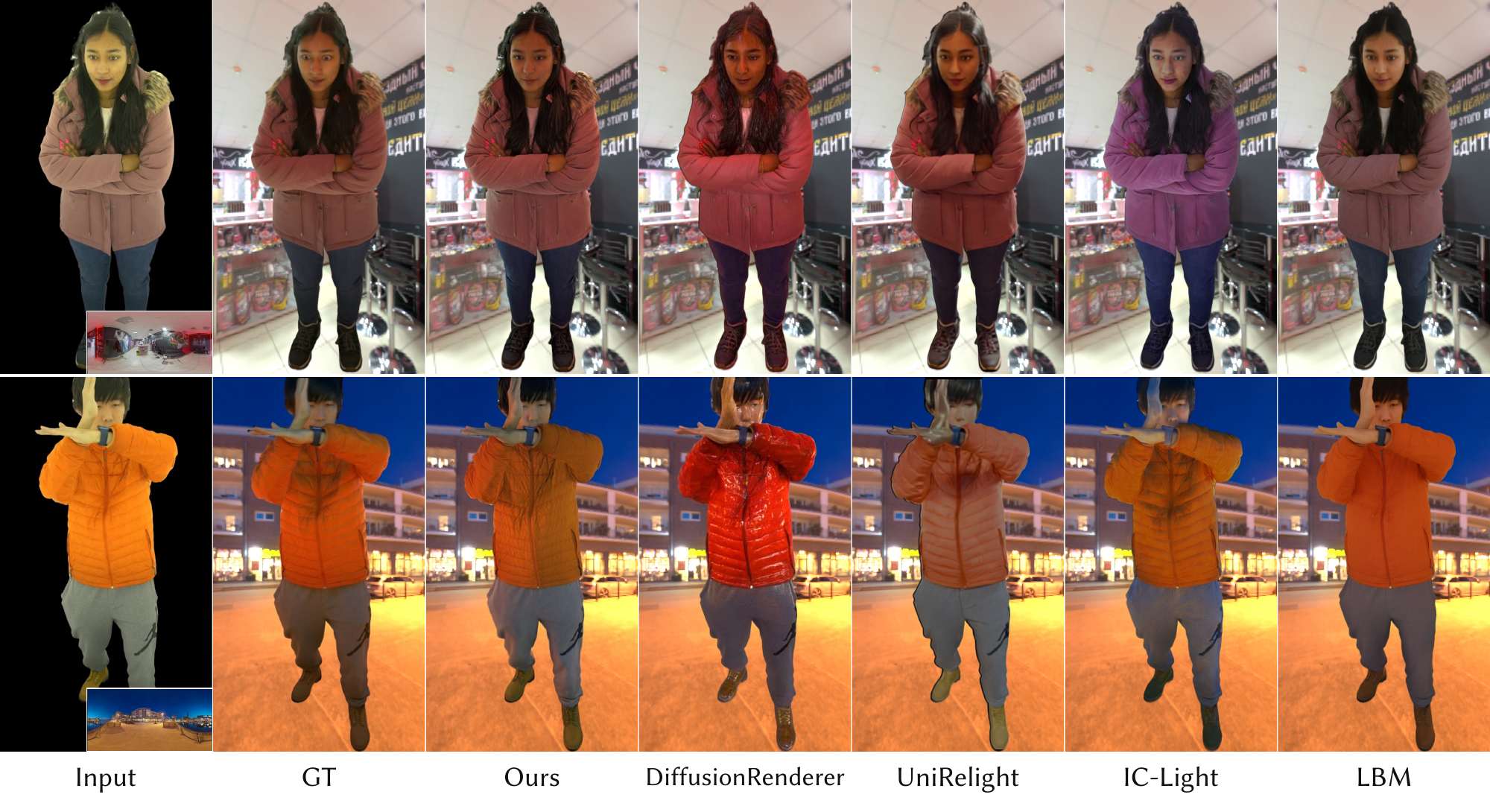}
    \caption{\textbf{Cross-dataset qualitative comparison on HumanOLAT~\cite{teufelgera2025HumanOLAT}.} All displayed method outputs are brightness-aligned to the corresponding ground truth for visualization.}
    \Description{HumanOLAT ground-truth images and brightness-aligned outputs from several relighting methods shown side by side.}
    \label{fig:supp_humanolat_comparison}
\end{figure*}

\begin{figure*}[t]
    \centering
    \vspace{-2mm}
    \begin{minipage}[t]{0.47\textwidth}
        \centering
        \includegraphics[width=\linewidth]{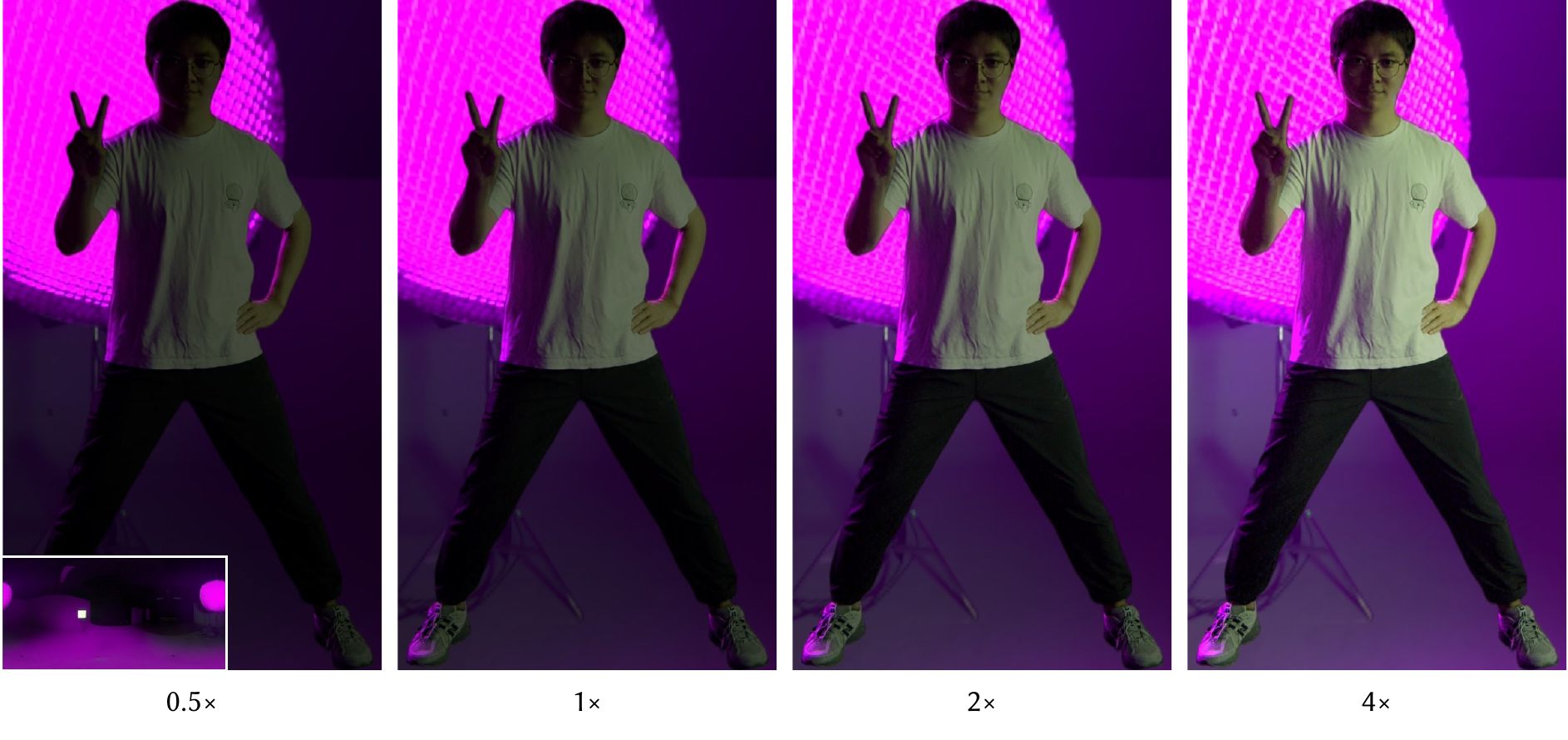}
        \captionof{figure}{\textbf{Relative intensity control under a fixed environment map.} Scaling the same environment map from $0.5\times$ to $4\times$ produces corresponding changes in the relit appearance.}
        \Description{Relit outputs generated with one environment map scaled from 0.5 times to 4 times intensity.}
        \label{fig:supp_intensity_control}
    \end{minipage}
    \hfill
    \begin{minipage}[t]{0.47\textwidth}
        \centering
        \includegraphics[width=\linewidth]{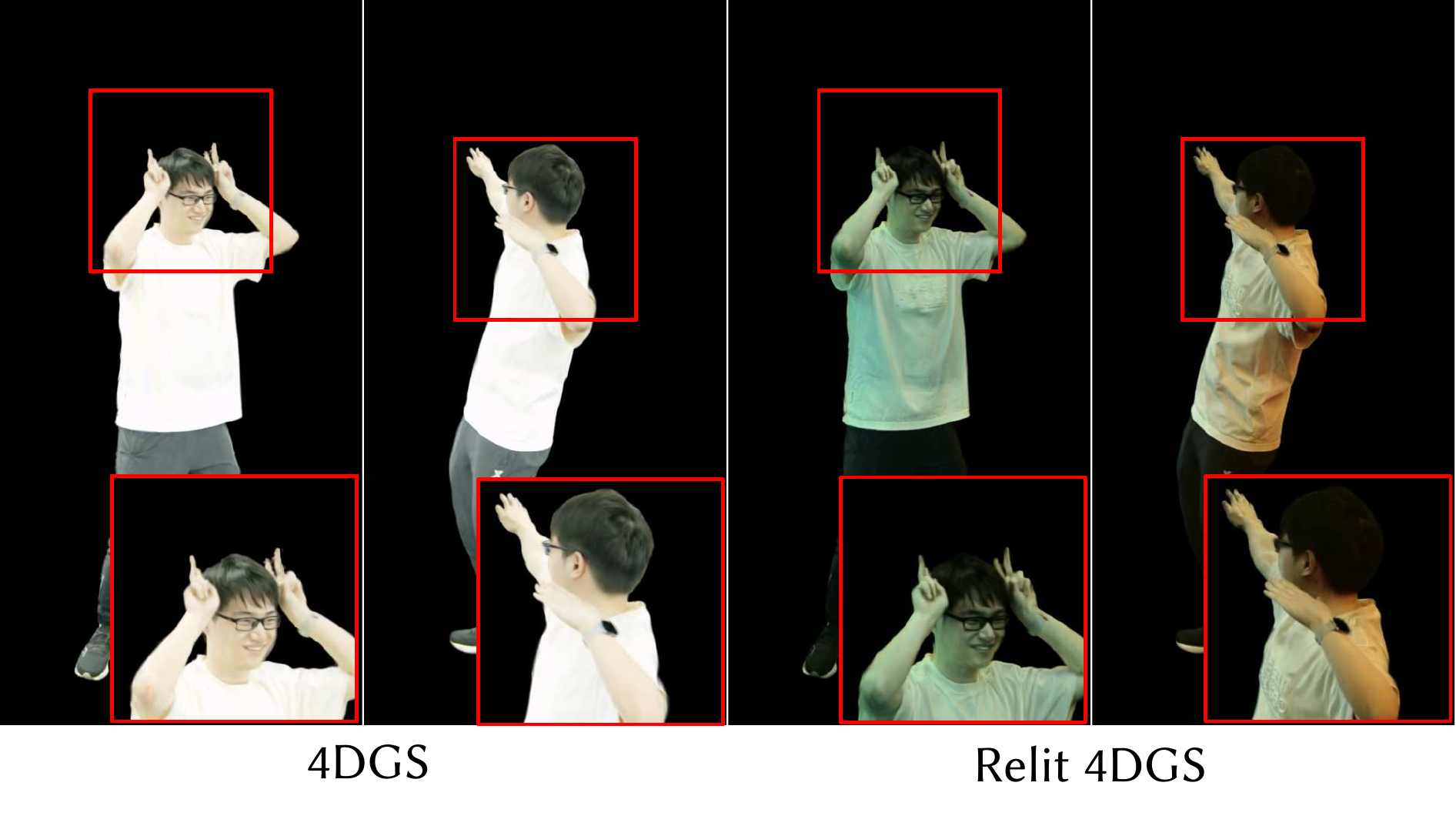}
        \captionof{figure}{\textbf{4DGS reconstruction from relit videos.} Left: Reconstruction from original flat-lit inputs. Right: Reconstruction from our relit outputs. The plausible novel view synthesis suggests that our method maintains sufficient implicit multi-view consistency to support downstream 3D/4D tasks.}
        \Description{Side-by-side novel-view renderings from 4D Gaussian Splatting models reconstructed using flat-lit and relit multi-view videos.}
        \label{fig:relit_gs}
    \end{minipage}
\end{figure*}

We provide additional qualitative results for the quality--cost tradeoff (Fig.~\ref{fig:supp_bases_comparison}), in-the-wild inputs (Fig.~\ref{fig:supp_wild_results}), long-sequence inference (Fig.~\ref{fig:supp_long_video}), cross-dataset evaluation (Fig.~\ref{fig:supp_humanolat_comparison}), and relative intensity control (Fig.~\ref{fig:supp_intensity_control}).

\section{Extended 4D Relighting}
\label{sec:supp_relighting}
To probe the implicit consistency discussed above, we optimize a LongVolcap 4D Gaussian Splatting (4DGS) representation from independently relit multi-view videos~\cite{xu2023easyvolcap, xu2024longvolcap}.
The plausible novel-view reconstruction in Fig.~\ref{fig:relit_gs} suggests that the generated videos retain sufficient implicit multi-view coherence for this downstream example.

\section{Dataset overview}
\label{sec:supp_overview}
\begin{figure*}
    \centering
    \includegraphics[width=0.95\textwidth]{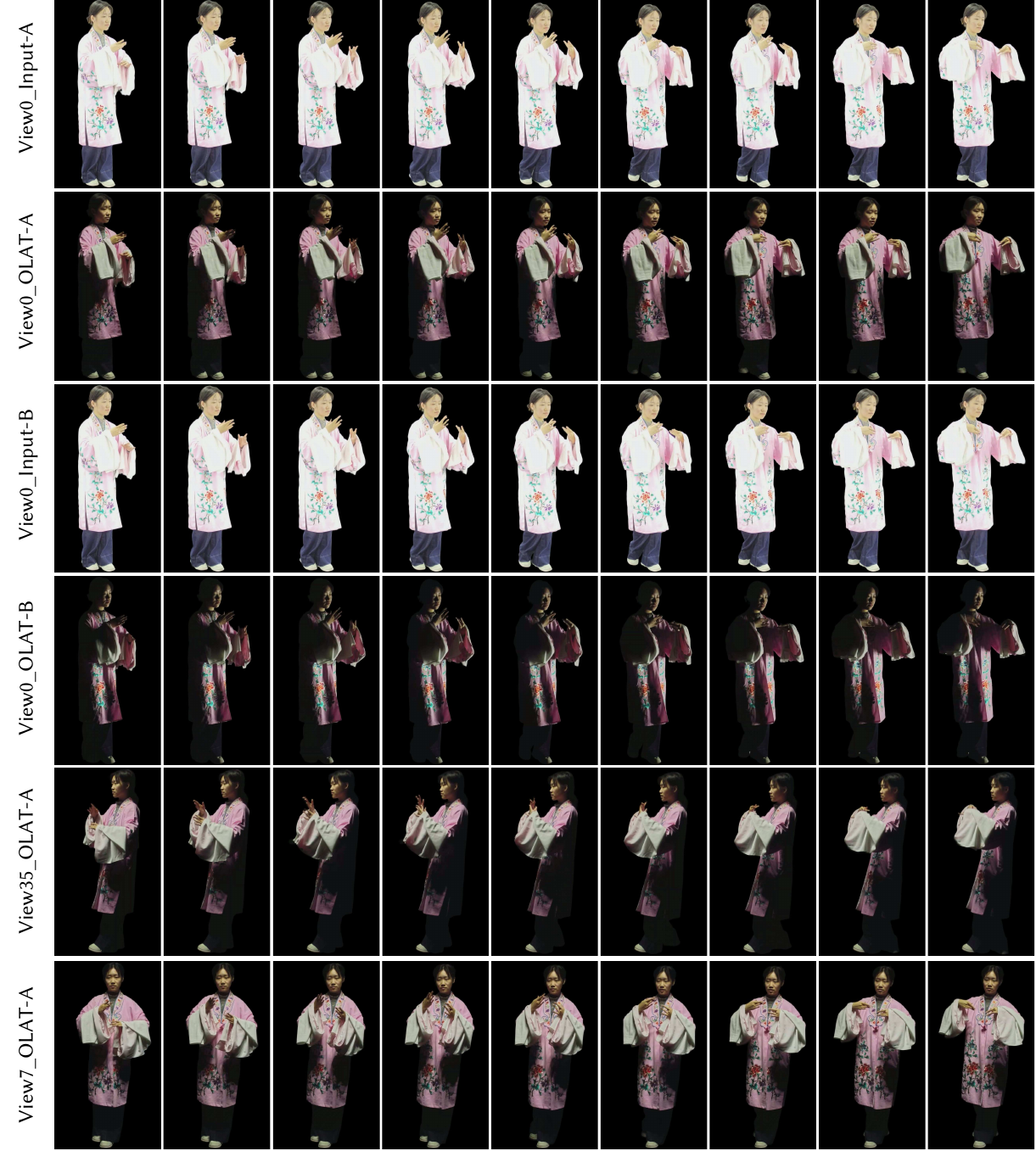}
    \caption{
        \textbf{Examples of training pairs from the \textit{Dynamic-subset}.}
        This figure shows the details of video training pairs in one sequence from several views.
    }
    \Description{Multi-view flat-lit and OLAT video-frame pairs from one dynamic capture sequence.}
    \label{fig:training_pairs}
\end{figure*}

\begin{figure*}
    \centering
    \includegraphics[width=0.8\textwidth]{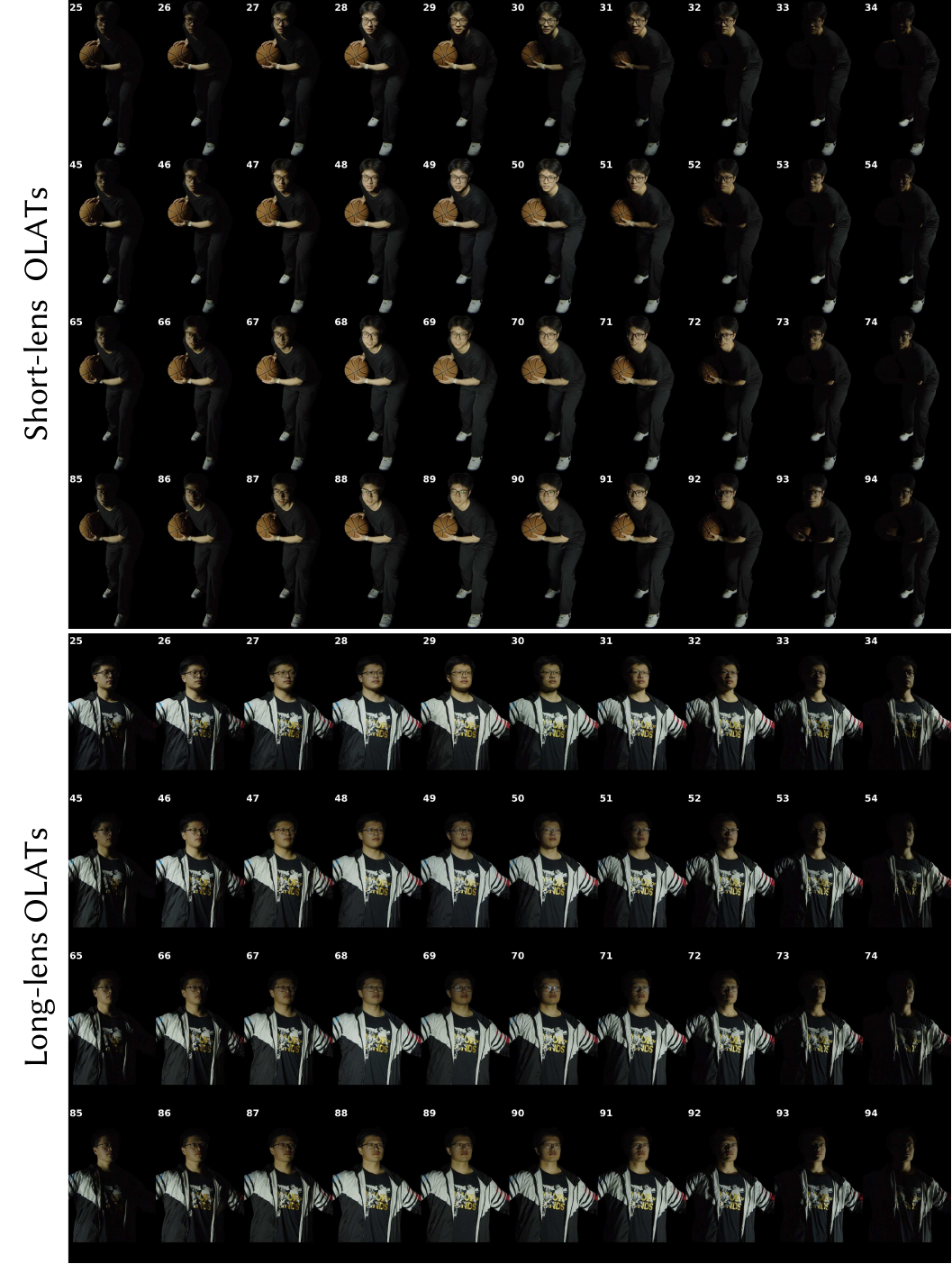}
    \caption{
            \textbf{OLAT images from short-lens and telephoto cameras in the \textit{Static-subset}.}
            The left-up corner number represent lighting direction index.
    }
    \Description{A grid of OLAT images from short-lens and telephoto cameras under several indexed lighting directions.}
    \label{fig:static-olats-overview}
\end{figure*}

\begin{figure*}
    \centering
    \includegraphics[width=0.92\textwidth]{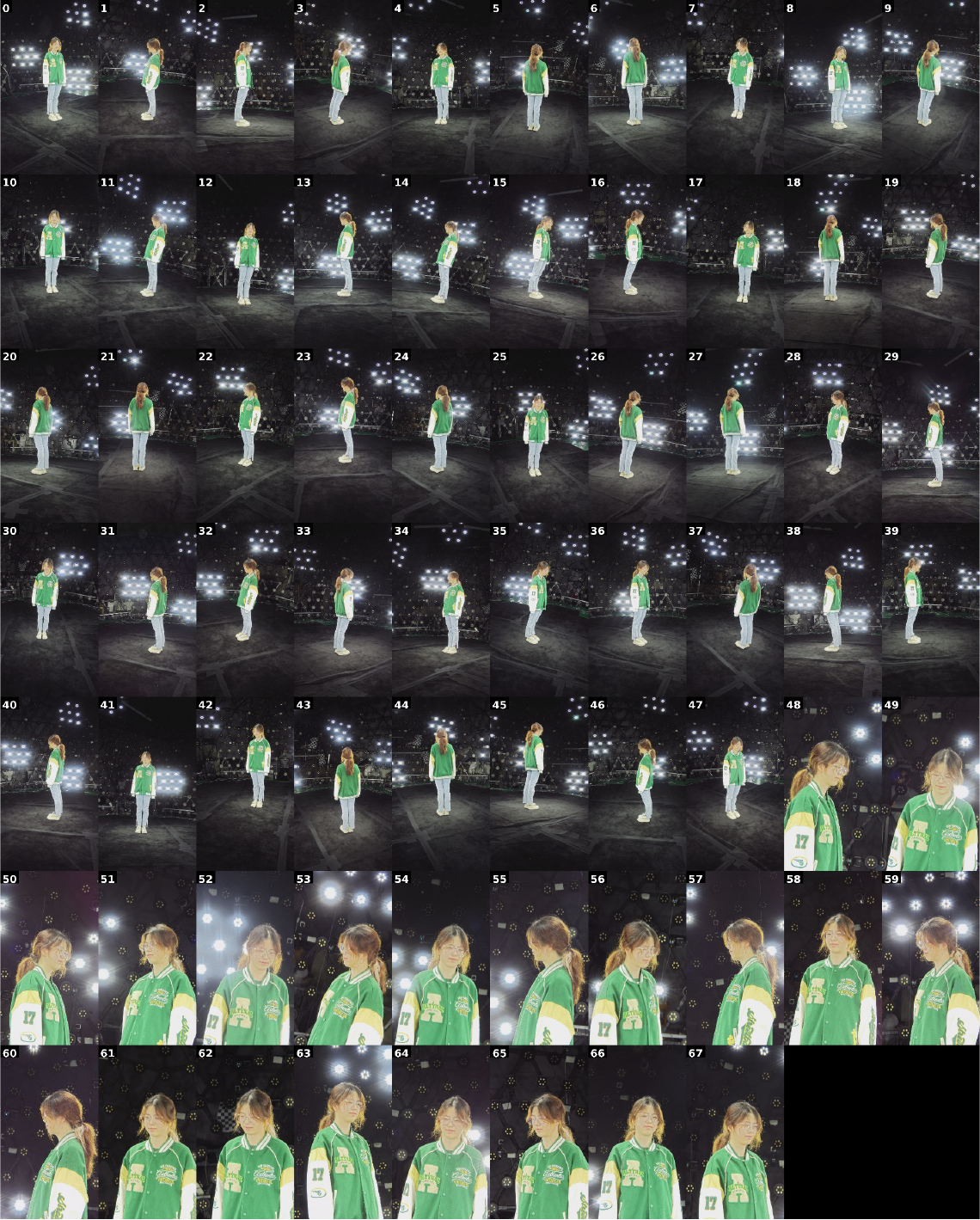}
    \caption{  
             Snapshots from all views of an example subject in the \textit{Dynamic-subset}.
    }
    \Description{Snapshots of one dynamic capture shown from all synchronized camera viewpoints.}
    \label{fig:dynamic-subset-allviews}
\end{figure*}

\begin{figure*}
    \centering
    \includegraphics[width=0.92\textwidth]{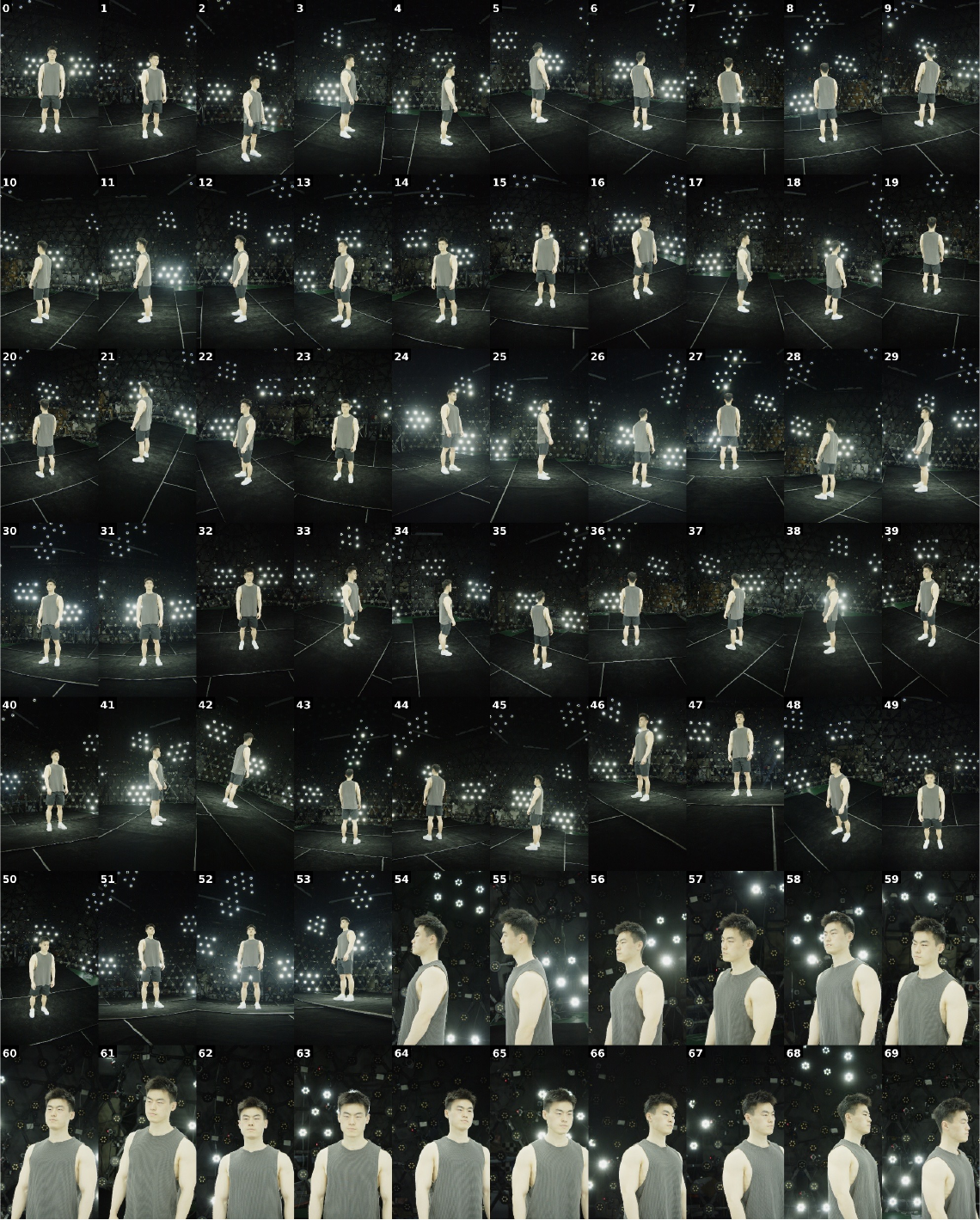}
    \caption{
            Snapshots from all views of an example subject in the \textit{Static-subset}.
    }
    \Description{Snapshots of one static capture shown from all synchronized camera viewpoints.}
    \label{fig:static-subset-allviews}
\end{figure*}

\begin{figure*}
    \centering
    \includegraphics[width=0.95\textwidth]{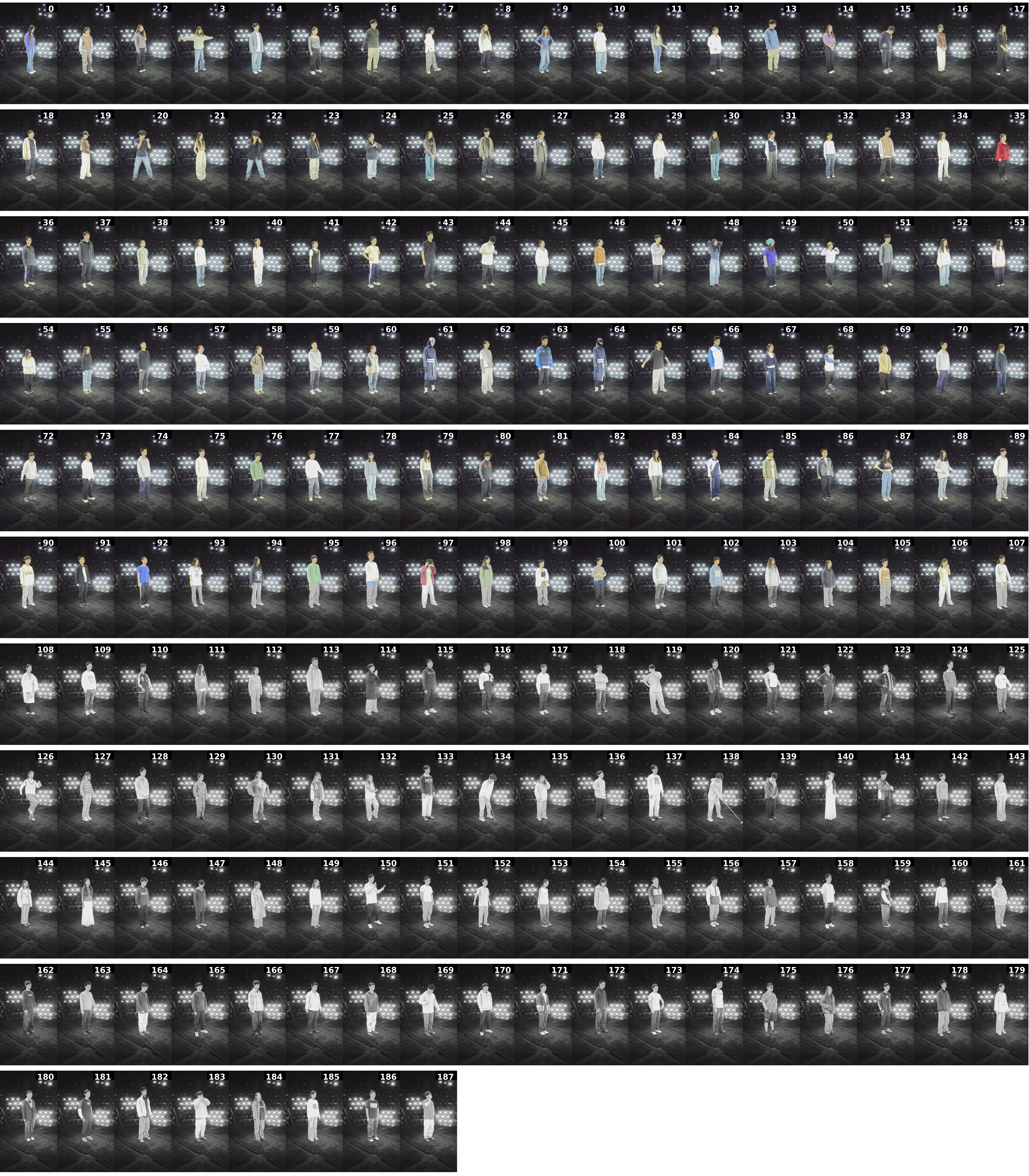}
    \caption{
        Overview of all identity-outfit combinations in the \textit{Dynamic-subset}.
    }
    \Description{Gallery of all identity-outfit combinations in the Dynamic-subset.}
    \label{fig:dynamic-subset-subjects-overview}
\end{figure*}

\begin{figure*}
    \centering
    \includegraphics[width=0.95\textwidth]{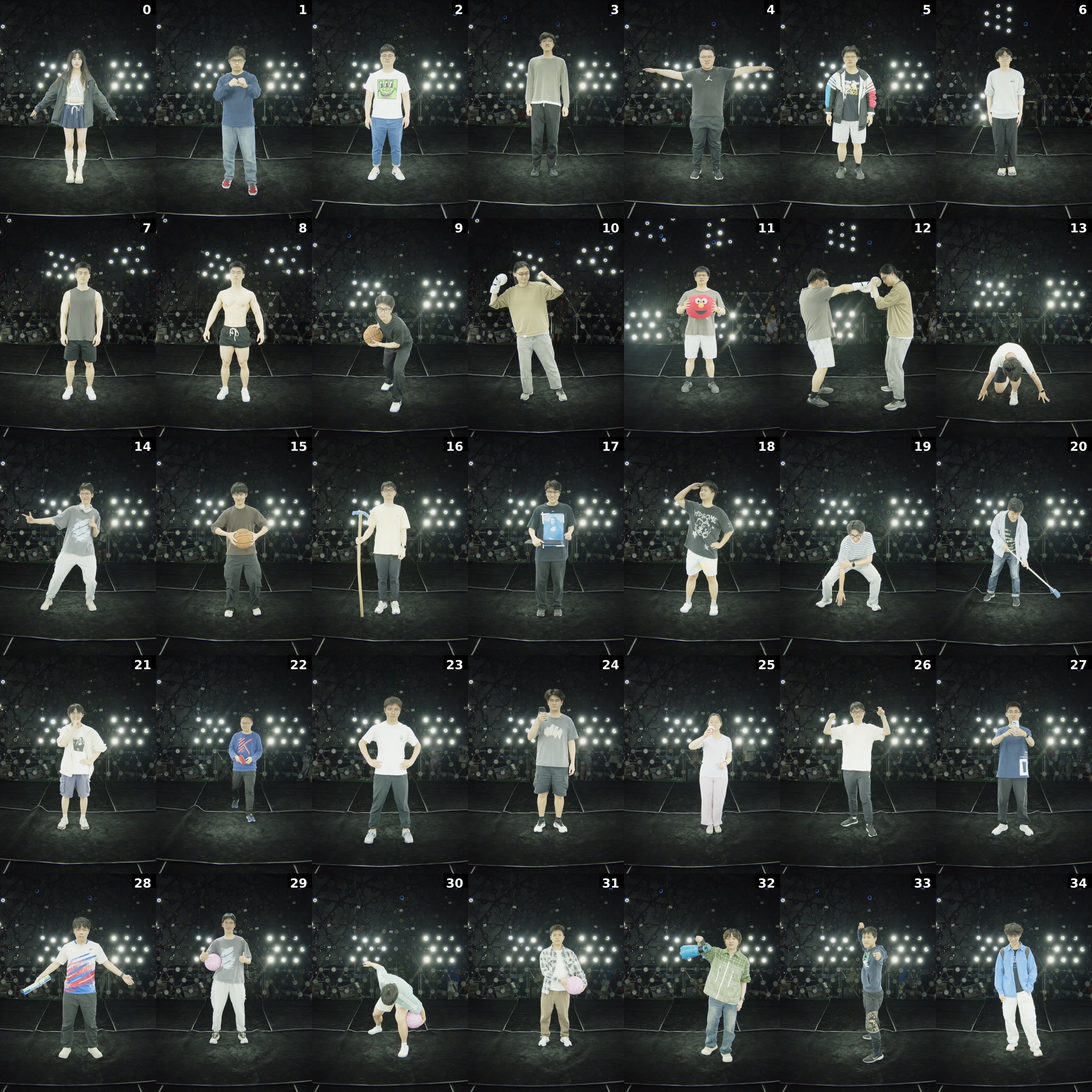}
    \caption{
        Overview of all identity-outfit combinations in the \textit{Static-subset}.
    }
    \Description{Gallery of all identity-outfit combinations in the Static-subset.}
    \label{fig:static-subset-subjects-overview}
\end{figure*}

We provide several visualizations to illustrate the composition and quality of our dataset. 
Specifically, Fig.~\ref{fig:training_pairs} and Fig.~\ref{fig:static-olats-overview} show the basic training units: multi-view video pairs for the \textit{Dynamic-subset} and OLAT images for the \textit{Static-subset}. 
The multi-view snapshots for individual subjects are displayed in Fig.~\ref{fig:dynamic-subset-allviews} (\textit{Dynamic-subset}) and Fig.~\ref{fig:static-subset-allviews} (\textit{Static-subset}). 
Furthermore, we present a gallery of all captured subjects in the \textit{Dynamic-subset} and the \textit{Static-subset} in Fig.~\ref{fig:dynamic-subset-subjects-overview} and Fig.~\ref{fig:static-subset-subjects-overview}.
\clearpage

\end{document}